\documentclass[runningheads,a4paper]{llncs}
\pdfoutput=1

\usepackage[T1]{fontenc}
\usepackage{graphicx}
\usepackage{amsmath,amssymb}
\usepackage{booktabs}
\usepackage{xcolor}   
\usepackage{placeins} 
\usepackage{hyperref}

\hypersetup{hidelinks}  

\begin{document}

\title{Reinforcement Learning with Verifiable Rewards for Small Search Agents}

\titlerunning{RLVR for Small Search Agents}
\author{Gaurisankar Jayadas\inst{1} \and
Aske Plaat\inst{1} \and
\'Alvaro Serra-G\'omez\inst{1} \and
Sandheep P\inst{2}}
\authorrunning{G. Jayadas et al.}
\institute{Leiden Institute of Advanced Computer Science,
Leiden University, Leiden, The Netherlands\\
\email{gaurisankarj1996@gmail.com}
\and
\email{sandheep.p01@gmail.com}}

\maketitle

\begin{abstract}
Reinforcement Learning with Verifiable Rewards (RLVR) performs well on problems
with clear rewards, such as mathematics and coding, but whether it also works
where the reward is less clear remains open. The reason-over-search recipe
applies RLVR to open-domain question answering, where retrieval grounds the
answer and a match against the reference supplies the reward. So far it has been
demonstrated on large models, and below one billion parameters only with
distillation from a larger teacher. We test the recipe on a small model. We
train Qwen3.5-0.8B with Group Relative Policy Optimization (GRPO) and an
interleaved Wikipedia-search tool on MuSiQue, varying only the reward across
three shapes over three seeds each, and we evaluate every checkpoint held-out on
a seven-benchmark question-answering suite. The recipe works: the best run
reaches 0.352 average exact match against a 0.092 untrained floor, a 3.8-fold
gain, with no distillation step in the training loop. The reward shape also
matters. The Search-R1-faithful exact-match-only reward is the worst of the
three at every seed at the matched training horizon, and it is worst even on
exact match, the metric it directly optimises. We conclude that the sparse
exact-match reward, RLVR's default in mathematics and code, is the wrong
starting point for models of this size. The reason-over-search setting can
supply a suitable reward for RLVR on small models, but small-model RLVR needs
its own reward-design study rather than a scaled-down copy of a large-model
recipe.

\end{abstract}

\section{Introduction}\label{sec:intro}

Large Language Models (LLMs) acquire broad competence from next-token
pretraining \cite{radford2019gpt2}, but their most striking recent gains on
multi-step problems come from Reinforcement Learning (RL) applied afterwards.
Two independent systems, o1 \cite{openai2024learning} and DeepSeek-R1
\cite{Guo_2025}, showed that large-scale RL over long \emph{reasoning traces},
the intermediate tokens a model generates before its final answer, keeps
improving accuracy as test-time computation grows. What makes this training
scale is that the reward is verifiable: for domains with checkable answers, such
as mathematics and code, a deterministic function of the generated answer and a
reference answer returns the reward directly, so no learned reward model is
needed, unlike Reinforcement Learning from Human Feedback
\cite{ouyang2022instructgpt}. This paradigm, Reinforcement Learning with
Verifiable Rewards (RLVR) \cite{lambert2024tulu3}, is at the core of most recent
progress on LLM reasoning benchmarks.

Such a model can still answer only from what its weights encode.
Retrieval-Augmented Generation (RAG) \cite{Lewis2020rag} lifts that bound by
concatenating documents fetched from an external corpus into the input. One
retrieve-then-read step, however, forms its single query from the question
alone, which is insufficient when the answer requires several linked facts:
there the second query can be formed only after the first retrieved passage has
been read, because that passage supplies the entity the second query must name.
One family of previous work addresses this by generating the queries and the
answer as one token sequence
\cite{react_yao_2023,press2023compositionality,ircot}: the model
generates tokens until it emits a search query, the retrieved passages are
concatenated into its context, and generation resumes, until it emits an answer.
A second family trains that same sequence end-to-end with RL against a single
reward computed from the final answer
\cite{Jin2025SearchR1TL,chen2025researchlearningreason,song2025r1searcher},
which removes the hand-written control flow. We call that second family, RL over
an interleaved search tool with an answer-match reward, the
\emph{reason-over-search} recipe. Its results are reported almost exclusively at
3B to 32B parameters, on previous model generations.

Smaller models are far cheaper, both to train and at inference time. They are
arguably the more suitable class for agentic systems
\cite{wang2024comprehensivesurveysmalllanguage,belcak2025smalllanguagemodels}.
But pretraining leaves them with a weaker \emph{prior}: their initial
next-token distribution places less probability mass on the behaviour the reward
is meant to select, so RL has less useful behaviour to reinforce and depends
more on how informative the reward signal is. The shape of that reward should
therefore become a more decisive design choice as the model gets smaller,
although a newer generation's pretraining and post-training partially offset the
deficit. The little evidence at sub-1B scale is cautionary: previous work finds
pure RL too unstable to bootstrap search behaviour from a cold start at 0.5B to
1B, and recovers it only by distilling from a larger teacher
\cite{kotoge2025dgpo}.

\paragraph{Problem statement and research questions.}
Whether the reason-over-search recipe transfers to a sub-1B model trained from
its own initialisation, and which reward shape works best at that scale, is
open. We ask: \emph{can RL with verifiable rewards over a search tool be made to
work on a sub-1B language model, and what reward shape is best?} Three research
questions decompose it. \textbf{RQ1 (feasibility and generalisation):} can Group
Relative Policy Optimization (GRPO) \cite{shao2024deepseekmath} with a search
tool train a sub-1B model to beat its untrained baseline on held-out multi-hop
question answering (QA)? \textbf{RQ2 (reward shape, accuracy):} which reward
shape gives the best held-out exact match? \textbf{RQ3 (reward shape,
stability):} which reward shape is the most stable across random seeds and
training horizon? We phrase the questions as ``sub-1B'' with Qwen3.5-0.8B as
the single instance; single-model scope is a limitation
(Section~\ref{sec:limitations}), not a claim about all sub-1B models.

\paragraph{Approach.}
We train Qwen3.5-0.8B \cite{qwen3.5} with GRPO on MuSiQue \cite{MuSiQue}, the
hardest multi-hop dataset in the reference papers' evaluations, and we evaluate
every checkpoint on a held-out seven-benchmark QA suite. To test whether the
reward shape changes held-out accuracy and seed stability, we train the model
under three reward shapes spanning a sparse-to-dense axis while keeping every
other component of the recipe fixed (Section~\ref{sec:setup}).

\paragraph{Contributions.}
\begin{itemize}
  \item \textbf{C1 (feasibility).} GRPO with a search tool trains Qwen3.5-0.8B
  from its own initialisation, with no distillation step inside the training
  loop, to clear its untrained floor by a wide margin on the held-out
  seven-benchmark suite, showing that the reason-over-search recipe transfers
  below one billion parameters, in contrast with the distillation-dependent
  earlier evidence at that scale \cite{kotoge2025dgpo}.
  \item \textbf{C2 (reward shape).} In a controlled ablation of three reward
  shapes across three seeds, the Search-R1-faithful exact-match-only reward is
  the worst shape at every seed at the matched horizon, worst even on the
  exact-match metric it directly optimises, while the dense token-F1 reward
  carries the gains.
\end{itemize}
Section~\ref{sec:mechanism} discusses which training-time measurements are, and
are not, consistent with a proposed explanation of C2.

\section{Related Work}\label{sec:related}

\subsection{RL for LLM reasoning}

RL entered LLM post-training as a way to align models with human intent, through
supervised fine-tuning, a learned preference reward model, and PPO
\cite{ouyang2022instructgpt,schulman2017ppo}. The shift since then is from
learned preference rewards to verifiable ones, where a deterministic function
scores the answer and the reward model disappears: DeepSeek-R1 \cite{Guo_2025}
incentivises reasoning traces by RL alone, with no supervised traces, and the
paradigm was crystallised as RLVR in open recipes such as Tulu 3
\cite{lambert2024tulu3}, over the step-by-step output form introduced by
Chain-of-Thought prompting \cite{wei2022cot}.

GRPO, the algorithm at the centre of this work, was introduced with the
DeepSeekMath recipe \cite{shao2024deepseekmath}; it removes PPO's learned value
critic, a network typically as large as the policy, and so roughly halves
training memory and compute (Section~\ref{sec:prelim} states the objective we
optimise). The headline RLVR
results sit in mathematics and code, where the check is near-unambiguous, and
almost entirely at 1.5B parameters and above. Our setting moves RLVR to
open-domain QA, where the signal is a noisy answer-match, and down to a sub-1B
policy.

\subsection{Retrieval-augmented reasoning with tool use}

Multi-hop questions exposed the limits of one-shot retrieval and motivated a
family of prompted methods that interleave retrieval with reasoning \cite{react_yao_2023,press2023compositionality,ircot,jiang2023flare,asai2024selfrag},
alongside Toolformer \cite{toolformer}, which learns tool calls self-supervised
(Self-Ask \cite{press2023compositionality} also contributes our Bamboogle
benchmark). All either hand-engineer the control flow in the prompt or learn it
through supervision rather than from an outcome reward.

The RL turn replaced that hand-engineering with a single outcome reward.
Search-R1 \cite{Jin2025SearchR1TL} trains a policy with PPO or GRPO to
interleave reasoning with multi-turn search calls over a frozen corpus, masking
retrieved tokens from the policy-gradient loss and rewarding only exact match on
the final answer. It is the reference recipe and the baseline we reproduce.
ReSearch \cite{chen2025researchlearningreason} trains the same behaviour with
GRPO and a token-F1 reward carrying a small partial-credit format floor, and
shows that training on a single multi-hop dataset generalises across benchmarks.
It is the recipe we port. R1-Searcher \cite{song2025r1searcher} instead uses a
two-stage outcome-based scheme with no supervised cold start, and reports a small
answer-reward ablation comparing exact match and token-F1, finding token-F1
better at 7B. A recent wave of work treats the single outcome
reward as too sparse and adds dense per-step credit, rewarding the necessity of
each search decision or the per-turn information gain
\cite{wu2025hiprag,IGPO,TIPS,IG-Search,Search-P1}.

Two gaps separate that work from ours. First, every RL-trained
reason-over-search result above sits between 3B and 32B parameters, almost all
on previous-generation model families; below 1B the recipe has been tested only
with distillation guidance \cite{kotoge2025dgpo}, not from the model's own
initialisation. Second, none of these recipes makes the reward shape itself the
controlled variable: Search-R1 fixes a sparse exact-match reward and leaves
richer reward terms to future work, ReSearch fixes a dense token-F1 reward with
a format floor, and R1-Searcher's comparison is a single-seed side ablation at
7B. We stay at the outcome level
deliberately, treating the dense per-step cluster as the foil a minimal recipe
is positioned against rather than as a component we adopt.

\subsection{Reward shaping, and whether tricks help}

GRPO has spawned a family of stabilising modifications
\cite{yu2025dapo,liu2025drgrpo,zheng2025gspo}. A counter-current questions
whether any is needed at small scale. JustRL \cite{justrl2025} shows that a
single-stage, essentially plain GRPO recipe with one fixed hyperparameter set,
reused across two 1.5B models, matches or beats elaborate pipelines at a
fraction of the compute. It also shows that standard additions can actively
hurt. Closest to this work, a study of how
to train deep-research agents \cite{xu2026howtotrain} runs a direct reward
ablation in the Search-R1 setting at 3B and above and finds reward shape
decisive, but in the opposite direction. There, a naive token-F1 reward
underperforms exact match through an \emph{answer-avoidance collapse}, in which
the policy stops emitting a final answer at all, and only an added action penalty
lifts F1 above EM.

JustRL's ``a minimal recipe can win'' is our design rationale: we vary only the
reward shape and keep plain GRPO otherwise, so any effect is attributable to the
reward rather than to a stack of interacting tricks. Our result also helps place
the apparent contradiction with \cite{xu2026howtotrain}. Token-F1 beats exact
match in two of the three reward ablations across scales (at 7B
\cite{song2025r1searcher} and at 0.8B here) and loses only where the
naive-token-F1 answer-avoidance collapse fires. Our own floorless F1-only reward
is likewise unprotected by a format term, yet does not trigger that collapse at
0.8B, so the contradiction tracks neither model scale nor the presence of a
format floor, but whether that failure mode fires; what predicts its firing
remains open.

\subsection{Efficiency in agentic RL, and small-model post-training}

Previous work shortens agentic rollouts by explicit intervention, either on the
training batch (rollout response recomposition \cite{rorecomp2025}) or on the
reward (tool-call budgets \cite{otc2025}, penalties on unnecessary search
\cite{wu2025hiprag}). A second line studies length as an emergent property, and
Concise Reasoning
\cite{fatemi2025concise} shows analytically that the default effect of RL is to
\emph{grow} response length, so explicit reduction requires a dedicated phase.
We report an effect that sits against both lines
(Section~\ref{sec:compression}). Most RLVR results, finally, sit at 1.5B
parameters and above
\cite{Guo_2025,justrl2025,zhan2025exgrpo,wang2025tina,wang2025oneshotrlvr}; the
evidence specific to small agentic-search models is the distillation-dependent
study already discussed \cite{kotoge2025dgpo}, against which our lever is
training from the model's own post-trained initialisation.

\section{Preliminaries}\label{sec:prelim}

\subsection{Problem setting}

The task is open-domain multi-hop question answering with an interleaved
retrieval tool. Given a question $x$, the policy $\pi_\theta$ generates a token
sequence. Whenever the generated text completes a search query $q$, generation
halts, a retriever returns the top-$k$ passages for $q$, those passages are
concatenated into the context as a tool-response block, and generation resumes
conditioned on everything in the context so far. The sequence terminates when
the model emits an answer block, at end-of-sequence, or when a turn or token
budget is exhausted. We call one such sequence a \emph{rollout} $o$. Retrieved
passages are inputs rather than model outputs, so they are excluded from the
training loss (Section~\ref{sec:method}). One scalar reward $r(o)\in[0,1]$ is
computed per rollout, as a deterministic function of the answer block it ends
with and the reference answers; no intermediate step is scored.

\subsection{Group Relative Policy Optimization}

GRPO \cite{shao2024deepseekmath} samples a group of $G$ rollouts per question and
maximises a clipped surrogate with an additive Kullback-Leibler (KL) penalty to a
frozen reference policy:
\begin{equation}
J(\theta) = \mathbb{E}\!\left[\frac{1}{G}\sum_{i=1}^{G}\min\!\Big(\rho_i A_i,\
\operatorname{clip}(\rho_i,\, 1-\varepsilon,\, 1+\varepsilon)\, A_i\Big)
- \beta\, D_{\mathrm{KL}}\!\left(\pi_\theta \,\|\, \pi_{\mathrm{ref}}\right)\right].
\end{equation}
Here $\pi_{\mathrm{ref}}$ is the frozen initialisation the KL term anchors to,
$\varepsilon$ the clip range and $\beta$ the KL weight (values in
Section~\ref{sec:setup}). Writing $o_i$ for the $i$-th rollout, $r_i = r(o_i)$
for its reward, and $\pi_{\mathrm{old}}$ for the policy that sampled the group,
which lags $\pi_\theta$ and is not $\pi_{\mathrm{ref}}$, the importance ratio and
the group-relative advantage are
\begin{equation}
\rho_i = \frac{\pi_\theta(o_i \mid x)}{\pi_{\mathrm{old}}(o_i \mid x)},
\qquad
A_i = \frac{r_i - \operatorname{mean}(\{r_j\}_{j=1}^{G})}{\operatorname{std}(\{r_j\}_{j=1}^{G})}.
\label{eq:advantage}
\end{equation}
In place of PPO's learned value critic, Equation~\ref{eq:advantage} baselines
each rollout against the mean and standard deviation of the rewards of the $G$
rollouts sampled for the same question. That equation is the one our central
result turns on. The update is driven by the \emph{spread} of rewards inside a
group, so a reward function that assigns most rollouts of a group the same score
drives $A_i$ toward zero, and with it that group's contribution to the gradient,
however many of those rollouts were wrong.

\section{Method}\label{sec:method}

\begin{figure}[tb]
\centering
\includegraphics[trim={12 315 477 195},clip,width=0.52\linewidth]{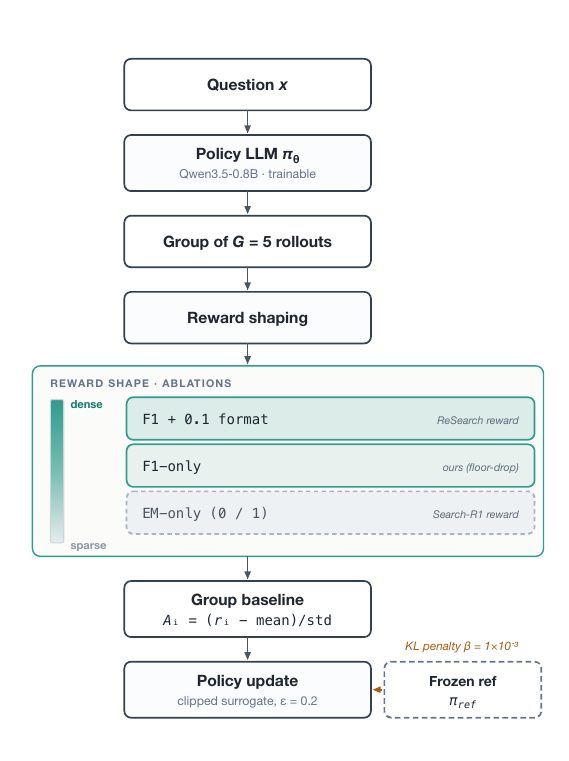}
\caption{The critic-free GRPO loop. A group of $G=5$ rollouts per question is
scored; the reward node branches into the three ablated shapes (F1 with a 0.1
format floor, F1-only, and sparse EM-only); the group mean and standard deviation
set the advantage baseline, with no value critic; the update is the clipped,
KL-regularised surrogate of Section~\ref{sec:prelim}. A worked
reason-search-read rollout, showing the interleaved tool calls and the loss
masking of retrieved spans, is in Appendix~\ref{app:rollout}.}
\label{fig:pipeline}
\end{figure}

Figure~\ref{fig:pipeline} overviews the training pipeline: the critic-free GRPO
loop of Section~\ref{sec:prelim}, with the reward node branching into the three
ablated shapes.

\subsection{Rollout construction and loss masking}

Three implementation choices define our instantiation of the rollout of
Section~\ref{sec:prelim}. First, the tool interface uses the Qwen3.5-native
nested-XML tool-call and tool-response tags rather than the invented string tags
of the reference recipes, so the format the policy must produce is
in-distribution for the model it is trained from. Second, retrieved tokens are
masked from the loss, as in both reference papers: the GRPO loss and its KL term
are computed only over model-generated tokens, implemented here by
message-role typing rather than by string offsets. Without this the policy would
be trained to predict corpus text it never chose. Third, the training overlay
imports the reward scorer, tool-call parser, and prompt template from the
evaluation package, so training and evaluation render byte-identical prompts
from one shared code object and no score difference between them can hide a
formatting drift.

\subsection{The three reward shapes}

The ablated variable is the function $r(o)$ of Section~\ref{sec:prelim}. All
three shapes read the last answer block and normalise it SQuAD-style
\cite{rajpurkar2016squad}. Exact match (EM) is all-or-nothing: 1 only if the
answer matches a gold answer word for word, 0 otherwise. Token-F1 gives partial
credit for word overlap (the harmonic mean of word precision and recall, taken
as the best over the gold-answer list). Writing $\hat a$ for the parsed answer,
$\mathrm{fmt}(o)$ for the format-validity check on a rollout $o$, and $Y$ for
the gold-answer set, the three shapes are
\begin{equation}
r_{\mathrm{EM}}(o)=\mathrm{EM}(\hat a,Y),\qquad
r_{\mathrm{F1}}(o)=\mathrm{F1}(\hat a,Y),
\end{equation}
both $0$ when $\hat a$ is absent, and the floored shape
\begin{equation}
r_{\mathrm{F1+fmt}}(o)=
\begin{cases}
\mathrm{F1}(\hat a,Y), & \mathrm{F1}(\hat a,Y)>0\\
0.1, & \mathrm{F1}(\hat a,Y)=0\ \wedge\ \mathrm{fmt}(o)\\
0, & \mathrm{F1}(\hat a,Y)=0\ \wedge\ \neg\,\mathrm{fmt}(o).
\end{cases}
\label{eq:rewards}
\end{equation}
The 0.1 is a floor that replaces F1, not an additive bonus: an F1 of 0.5 yields
a reward of 0.5.

The three shapes are chosen to span a reward-density axis from sparse to dense
while keeping two of them anchored to published recipes. EM-only is Search-R1's
outcome reward \cite{Jin2025SearchR1TL}, at the sparse end: a rollout scores 0
unless it is exactly right. F1+format is ReSearch's reward
\cite{chen2025researchlearningreason} at the dense end, except that we drop its
\verb|\boxed{}| gate, making it value-equivalent rather than byte-identical.
F1-only is our own floor-drop ablation, and isolates the contribution of the 0.1
floor: it was motivated by an earlier pilot of ours on a 0.6B model, where the
floor appeared to mask the tool-use signal, and it tests whether that finding
carries to 0.8B. Under
Equation~\ref{eq:advantage}, moving along this axis changes how often the $G$
rollouts of a group receive distinguishable scores, which is why we expect the
axis to matter at all.

\section{Experimental Setup}\label{sec:setup}

Nine runs cross the three reward shapes of Section~\ref{sec:method} with three
random seeds. Because the model, training data, prompt, retriever, and optimiser
settings are identical across all nine, any difference that exceeds seed
variance is attributable to the reward shape.

\subsection{Model, data, and retrieval}

The policy is Qwen3.5-0.8B \cite{qwen3.5}, a post-trained hybrid model that can
generate a reasoning trace before answering or answer directly; we avoid calling
it ``distilled'', since no primary source documents how the small variants were
produced. The prompt was fixed by its own two-step ablation before any reward run
(a minimal user message in place of a verbose system-protocol default lifted
untrained EM 6.6-fold; a 12-variant selection then chose the terse form, worth
roughly a further 1.7-fold), so all nine runs share one prompt-optimised
baseline.

We train on MuSiQue \cite{MuSiQue} only, a 2-to-4-hop dataset and the hardest
benchmark in the reference papers' evaluations; the single-dataset corpus is by
design, so that generalisation is measured strictly on held-out data.
The held-out suite comprises seven benchmarks: the single-hop Natural Questions
\cite{natural_questions_nq}, TriviaQA \cite{TriviaQA}, and PopQA \cite{PopQA};
the 2-hop HotpotQA \cite{HotpotQA} and Bamboogle
\cite{press2023compositionality}; and the 2-to-4-hop 2WikiMultiHopQA
\cite{2WikiMultiHopQA} and MuSiQue, totalling 51,713 rows per checkpoint. Six of
the seven are therefore strictly out-of-distribution.

Retrieval uses the Wikipedia-2018 corpus \cite{karpukhin2020dpr} with the
E5-base-v2 encoder \cite{wang2022e5} through FlashRAG \cite{flashrag}, over an
IVF4096-SQ8 quantised index \cite{douze2024faiss}, for both training and
evaluation. The reference papers use an exact flat inner-product index; we use it
nowhere, because under training rollout load it times out, and it needs roughly
65\,GB of host RAM against about 16\,GB for the quantised index, whose recall@10
we observed at 96 to 99\% of the flat index during bring-up (an engineering
observation, not a logged experiment). Our
Search-R1 3B reproduction keeps
that pipeline's top-3 default while our own runs use top-5, the one
retrieval-depth difference between them.

\subsection{Training configuration}

We use $G=5$ rollouts per question, clip $\varepsilon=0.2$, and KL weight
$\beta=10^{-3}$ with the k3 estimator. Generation runs until a stop string or
end-of-sequence; on a tool call the retriever service is queried at top-5, each
retrieved chunk is injected as its own tool-response block, and generation
resumes. A rollout ends on an answer, on end-of-sequence, or at a cap of 10
turns and 8192 tokens. The turn cap rarely binds in training; the token cap
binds often early and is central to Section~\ref{sec:compression}. The pilot
study ran on verl \cite{sheng2024hybridflow}; the main study moved to NeMo-RL
\cite{nemo_rl} because verl does not support Qwen3.5, a port forced by tooling
rather than preference. Training ran on a four-GPU RTX PRO 6000 Blackwell node,
at roughly \$120 to \$140 of rented compute per full-epoch run, which is the
budget that fixed the horizons below. Appendix~\ref{app:hparams} lists the full
shared configuration.

\subsection{Horizon, seeds, and evaluation protocol}

The training horizon is roughly one epoch, about 312 steps on MuSiQue, with
three seeds per reward shape (42, 43, 44). Seeds control data ordering and
rollout sampling, so the three shapes at a given seed share a data order: the
comparison is paired within seed and independent across seeds. We cap each seed
at 180, 230, and 312 steps respectively (checkpoints saved every 10 steps, to
step 310; we call each 10-step evaluation interval a \emph{cadence}). Realised
lengths differ, and three legs that ran past their cap are truncated to it, with
no effect on the ranking (Appendix~\ref{app:hparams}). Runs were length-bounded
by the compute grant, not stopped on any performance criterion. We therefore
report cross-seed claims at the
\textbf{matched horizon} (the best checkpoint at step $\le 180$, the largest
horizon common to all nine legs), since best-of-run is length-biased, alongside
the explicitly flagged \textbf{seed horizon} (best checkpoint within each
seed's own length).

Evaluation uses greedy decoding, matching the convention in the official code of
both reference recipes \cite{Jin2025SearchR1TL,chen2025researchlearningreason}
(neither paper states its evaluation decoding in text). The primary metrics are EM and token-F1, with
substring-cover accuracy (ACC, 1 if a normalised gold answer appears as a
substring of the normalised prediction) as a third, more lenient check. All
three are computed on the extracted answer span after the same normalisation,
and all three are defined independently of the training rewards that happen to
reuse two of them. Each checkpoint is evaluated over all
51,713 rows using the byte-aligned prompt shared with training, and every
reported figure is the unweighted macro-average over the seven benchmarks.
Each reported cell is the best checkpoint within the
stated horizon, selected on the same evaluation suite it is reported on; we
had no separate validation split, and Section~\ref{sec:limitations} discusses
what that costs.

\section{Results}\label{sec:results}

We evaluate nine GRPO runs (three reward shapes times three seeds), scoring
every checkpoint on the 51,713-row held-out suite; every EM number is the
seven-benchmark average, and the untrained floor is 0.092.

\subsection{Feasibility below one billion parameters (C1)}

Every trained run clears the 0.092 untrained floor by a wide margin, at every
seed and under both horizon rules (Table~\ref{tab:matched}).
The best run reaches 0.352 average EM (F1-only, seed 44, step 310), an absolute
gain of $+0.260$ and a 3.8-fold improvement; Table~\ref{tab:app-perbench} in
Appendix~\ref{app:tables} breaks each of the nine runs out over the seven
benchmarks. Because training is on MuSiQue only, six of those seven are strictly
out-of-distribution, so the lift on them is generalisation rather than
memorisation. This reproduces the ReSearch
\cite{chen2025researchlearningreason} headline that a model trained on MuSiQue
alone generalises across benchmarks, and extends it in two ways: to a sub-1B
model, and to the three single-hop benchmarks that ReSearch never evaluated. The
best run was still climbing at the budget horizon
(Figure~\ref{fig:app-s44}, Appendix~\ref{app:perseed}), so 0.352 is a
conservative lower bound rather than a converged ceiling.

\paragraph{Evaluation-pipeline validation.}
That lift is only interpretable if the evaluation pipeline is itself sound, so we
validate it against published numbers: run on our pipeline, the Search-R1
Qwen2.5-3B GRPO checkpoints reproduce within $\pm 2.5$ percentage points of the
values reported for them \cite{Jin2025SearchR1TL} (base $-2.4$, instruct $+0.5$).
Those same reproduction runs also place our 0.8B endpoint next to a
previous-generation 3B one. Since the two differ in model, generation, corpus,
reward, and recipe, we report that comparison as context in
Appendix~\ref{app:crossgen} rather than as a result.

\subsection{The reward-shape ablation (C2)}

Table~\ref{tab:matched} gives the central result. At the matched horizon (its
upper panel),
EM-only is the worst shape at all three seeds (0.301, 0.264, 0.249) and has the
widest seed range (0.052): it is both lowest and least stable.
Figure~\ref{fig:heldout-bar} in Appendix~\ref{app:metric} plots the same nine
cells as one bar per seed.

\begin{table}[t]
\centering
\caption{Held-out average EM by reward shape under both horizon rules. Upper
panel: the matched horizon (best checkpoint at step $\le 180$), where EM-only is
worst at every seed. Lower panel: each seed's own horizon (seed 42 at $\le 180$,
43 at $\le 230$, 44 at $\le 312$), where F1-only owns the single best run and
F1+format has the tightest spread.}
\label{tab:matched}
\begin{tabular}{lrrrrr}
\toprule
Reward shape & seed 42 & seed 43 & seed 44 & mean & range \\
\midrule
\multicolumn{6}{l}{\emph{Matched horizon (step $\le 180$)}} \\
F1+format & 0.322 & 0.292 & 0.308 & \textbf{0.307} & \textbf{0.030} \\
F1-only   & 0.313 & 0.283 & 0.318 & 0.305 & 0.035 \\
EM-only   & 0.301 & 0.264 & 0.249 & 0.271 & 0.052 \\
\midrule
\multicolumn{6}{l}{\emph{Each seed's own horizon}} \\
F1-only   & 0.313 & 0.291 & \textbf{0.352} & \textbf{0.318} & 0.061 \\
F1+format & 0.322 & 0.313 & 0.309 & 0.315 & \textbf{0.013} \\
EM-only   & 0.301 & 0.264 & 0.318 & 0.294 & 0.054 \\
\bottomrule
\end{tabular}
\end{table}

Three seeds are enough for an exact test of the headline ranking. Under the null
that reward shape has no effect, the three shapes are exchangeable within each
seed, so the probability that a pre-specified shape ranks last at all three
independent seeds is $(1/3)^3 = 1/27 \approx 0.037$. EM-only is pre-specified in
the required sense: it is the Search-R1-faithful arm the ablation was designed
to test, fixed before any held-out number was read, and the 0.6B pilot in fact
predicted a different loser (the floored shape). The direction, however, was not
pre-registered, so we report two conservative variants alongside: the two-sided
read on the same arm is $2/27 \approx 0.074$, and without pre-specifying any
shape, the probability that \emph{some} shape ranks last at all three seeds is
$3/27 \approx 0.111$. The token-F1 and accuracy orderings below score the same
predictions and are not independent evidence. All three readings are conditional on the matched
horizon: at each seed's own horizon EM-only ranks last at two of three seeds
(lower panel of Table~\ref{tab:matched}). Appendix~\ref{app:perseed} plots the
per-seed held-out trajectories behind both panels.

The one place EM-only looks competitive is at seed 44's full horizon (0.318,
edging out F1+format's 0.309 in the lower panel of Table~\ref{tab:matched}), but that leg ran to step
310, and at the matched step $\le 180$ horizon it scores only 0.249. We read
that as a length effect; Section~\ref{sec:limitations} gives the slow-start
reading we cannot rule out.

\paragraph{The ordering is metric-robust.}
A natural objection is that EM-only is being judged on a metric that flatters
its rivals. It is not. Figure~\ref{fig:3metric} (Appendix~\ref{app:metric})
reads all three metrics at each run's best-EM checkpoint at the matched horizon.
EM-only is lowest on all three (EM 0.271, F1 0.340, ACC 0.300) against F1-only
(0.305 / 0.393 / 0.370) and F1+format (0.307 / 0.389 / 0.353), and all three
shapes clear every metric's untrained floor (EM 0.092, F1 0.123, ACC 0.146). Its
deficit on token-F1 is in fact \emph{larger} than on EM, and it is lowest even
on substring accuracy, the only metric that is no shape's training objective, so
the ranking is a property of the policy rather than of the scoring metric
(Appendix~\ref{app:tables} tabulates per-seed token-F1 and accuracy cells under
their own selection rules).

\paragraph{Which F1 shape wins is seed-dependent.}
In both panels of Table~\ref{tab:matched}, F1+format wins seeds 42 and 43
while F1-only wins seed 44. We report this as seed-dependent
rather than collapsing it: the F1+format minus F1-only gap at the matched
horizon is $+0.9$, $+0.9$, and $-1.0$ percentage points, a mean of $+0.3$ whose
sign flips, well inside seed noise. The one robust matched-horizon ordering is
EM-only last: both F1 shapes beat it at every seed, F1-only by $+1.2$, $+1.9$,
and $+6.9$ points. Across both panels, F1+format has the tightest seed range
and F1-only the highest ceiling (0.352, at the cost of the full 310-step leg),
which gives a practical trade-off: prefer F1+format for low-variance behaviour
across seeds, F1-only for the best attainable endpoint. At $n=3$ the stability
half of that advice rests on a three-point spread and the ceiling half on the
single full-epoch seed. Cadence-to-cadence trajectory stability
(Figures~\ref{fig:app-s42} to~\ref{fig:app-s44}) also ranks mostly by seed
rather than by shape, so F1+format's seed-robustness should not be over-read as
the calmest trajectory.

\paragraph{Where the format floor's seed-42 advantage comes from.}
\emph{Observed}, over the held-out rollouts that produce the seed-42 column of
Table~\ref{tab:matched}, at the two runs' matched-horizon best checkpoints:
F1+format's $+0.9$ percentage-point EM lift over F1-only comes with a $+8.9$ point higher
valid-answer rate (89.3\% to 98.2\%) and a $+5.5$ point higher answer-token
retrieval-grounding, the fraction of answer tokens appearing in the retrieved
passages (0.830 to 0.885), both far larger than the EM difference they
accompany. \emph{Hypothesised}: the format gate acts upstream of the exact
match, pulling the policy toward a well-formed, retrieval-grounded answer, only
a fraction of which converts into an additional exact match. We intervened on
neither intermediate quantity, so we cannot test that ordering here, and the
lift is within seed noise and reverses at seed 44. What is established is
narrower: at seed 42 the floor did not mask the tool-use signal, the opposite of
what our 0.6B pilot suggested.

\subsection{A proposed mechanism for the exact-match deficit}
\label{sec:mechanism}

This subsection continues the discussion of C2, and is a hypothesis with
supporting evidence rather than a further result. \emph{The hypothesis} is that
the deficit is produced by the group-relative advantage of
Equation~\ref{eq:advantage}. At this scale exact matches are rare, so a sparse
0/1 reward gives most of a group's $G$ rollouts the same score of 0; the
within-group spread, and with it the advantage, goes to zero, and the group
contributes almost nothing to the update even though nearly all of its rollouts
were wrong. A dense token-F1 reward separates those same rollouts by partial
overlap and keeps the spread informative.

Directly testing this would require the per-group reward variance, which we did
not log. What we can report is that two training-time quantities behave as the
hypothesis predicts, computed over all nine runs and plotted for seed 42 in
Figure~\ref{fig:app-health} (Appendix~\ref{app:health}). EM-only has the lowest
gradient norm of the three shapes at every seed. Read to each seed's own horizon
it also fails to sharpen its policy entropy at two of the three seeds, ending
near its cold-start value at seeds 42 and 43 while entropy declines for five of
the six F1 legs. That entropy gap is a full-leg read: within step 180 it is
present at seed 42 alone. Neither observation is decisive on
its own, since other differences between the shapes could produce them. What
they do rule out is a competing explanation: gradient norm stays bounded at
every seed and none of the nine data-parallel runs diverged, so EM-only's
deficit is a weak-signal effect rather than a training instability.

Figure~\ref{fig:dyn} (Appendix~\ref{app:dyn}) shows the compact cross-seed read
of the behaviour that results. Averaged across seeds at the matched horizon, the
F1 shapes lean (falling tool calls and response length) while their reward
climbs, whereas EM-only stays highest on tool calls. Reward levels are not
comparable across shapes (Section~\ref{sec:ceiling}), so only the within-shape
trends and the behavioural panels carry weight here.

The reward panel of Figure~\ref{fig:dyn} also separates three trajectory
archetypes, visible per seed in Figures~\ref{fig:app-dyn42}
to~\ref{fig:app-dyn44}. F1-only shows a slow ramp and a smooth climb with no
catastrophic collapse. F1+format shows a medium-fast ramp and the highest peak,
and is also the steadiest of the three in training reward (per-step coefficient
of variation 0.12 to 0.15 from step 50, against 0.24 to 0.26 for EM-only).
EM-only shows the
fastest early ramp and then a flat, low plateau. At seed 43 its
held-out EM peaks early and then slips (Figure~\ref{fig:app-s43}); at seed 44 it
instead recovers late (Figure~\ref{fig:app-s44}), which we read as a length
effect but which a slow-start reading would also predict.

If the hypothesis holds, the deficit should appear at evaluation time as
degraded rollout-level behaviour rather than as uniformly worse answers, and
that is what we find. Pooled over the three seeds on the four multi-hop
benchmarks at each run's best-EM checkpoint (67{,}569 rollouts per shape,
Figure~\ref{fig:app-searchcount}), completed rollouts trace the same inverted-U,
peaking at two searches for every shape, and at that peak EM-only is the
strongest answerer, not the weakest (0.386 against F1-only's 0.378 and
F1+format's 0.341).
What changes is the abort rate, 22.8\% for EM-only against 12.7\% for F1-only
and 5.0\% for F1+format, and an aborted rollout scores zero by construction, so
EM-only's multi-hop held-out deficit is better read as a failure to terminate
than as worse answers. Conditioning on completion selects a different fraction of
rollouts per shape, so we read these curves as descriptive
(Appendix~\ref{app:termination}).

\subsection{Emergent compression, and its reward dependence}
\label{sec:compression}

Reward density has a second, practical consequence. Under an unmodified outcome
reward carrying no length term, all three shapes shorten their rollouts early,
but only the F1 shapes stay short. The separation that holds at every seed is in
the length-clip ratio, the fraction of rollouts truncated at the generation cap.
Both F1 shapes drive it to about 1\% by the end of every leg, while EM-only
still leaves 9.7\%, 14.1\% and 35.6\% at seeds 42, 43 and 44. Over a full epoch
the gap widens into raw budget: at seed 44 (Table~\ref{tab:app-compression},
Appendix~\ref{app:compression}) F1-only cuts tool calls from 6.6 to 3.4 and
total tokens from 6435 to 3694, and F1+format similarly, while EM-only leans
mid-horizon and then re-expands to end near its cold start, at 6.2 calls and
6031 tokens.

We hypothesise that the driver is the 8192-token generation cap acting as an
implicit length penalty: a rollout that runs to the cap never closes its answer
and scores 0, so shortening rollouts raises expected reward without any length
term in $r(o)$. Three measurements over the training logs are consistent with
it. First, hitting the cap and failing to produce a scorable answer are almost
the same event: the length-clip ratio and the no-valid-answer rate correlate at
Pearson $+0.98$ to $+0.99$ in every run.
Second, reward anti-correlates with the clip ratio for the F1 shapes ($-0.44$ to
$-0.52$ across all six legs). Third, at the full-epoch seed-44 leg that
anti-correlation is much weaker for EM-only ($-0.20$ against F1-only's $-0.44$),
as it should be if the pressure is what EM-only lacks, since under EM-only a
finished-but-wrong rollout scores 0 exactly like a truncated one. Across seeds,
though, EM-only spans $-0.20$ to $-0.50$, so this third measurement is suggestive
rather than decisive (Appendix~\ref{app:compression}). The evidence is correlational, and one alternative is
untested: Qwen3.5's post-training may already favour concise tool use, and we
did not run the base-model control that would separate the model prior from the
cap.

Compression is a cost property, not an accuracy lever. Across all the
checkpoint evaluations behind Table~\ref{tab:app-perbench}, held-out EM
correlates \emph{positively} with tool-call count in 7 of the 9 runs, so leanness
itself buys no accuracy. The win is against the starting point rather than across
checkpoints: at seed 44 F1-only ends the epoch at 3.4 tool calls and 3694 tokens
per sample, roughly half its cold start, while held-out average EM rises from the
0.092 untrained floor to 0.352. That is a deployment-cost win, not ``leaner means
more accurate''. It runs
opposite to the growing budget reported at larger scale
\cite{Jin2025SearchR1TL,chen2025researchlearningreason} and is driven by fewer
searches rather than shorter reasoning.

\subsection{A higher training reward is not more exact matches}\label{sec:ceiling}

A caution for anyone comparing training curves across shapes: a shape can raise
the optimisation scalar without producing more correct answers. The numbers in
this subsection come from the training logs behind Figures~\ref{fig:app-dyn42}
to~\ref{fig:app-dyn44}, not from the held-out suite. At seed 44, F1+format posts
the highest run-high reward (0.358) but the lowest run-high training-time
exact-match rate (20.5\%), against F1-only (0.314, 21.9\%) and EM-only (0.216,
21.6\%), because the 0.1 floor pays well-formed wrong answers. At the first
cadence the three shapes are tied on exact match (4.5 to 4.8\%) while their
scalars already differ 2.5-fold (EM-only 0.045, F1-only 0.072, F1+format 0.113)
from formatting alone. Training reward is therefore not
comparable across shapes, which is why we rank on held-out EM throughout. All
three shapes also peak at a training-time exact-match rate of 20 to 22\%, a
cross-shape spread of only 1.4 percentage points, which we read as a ceiling on
the training signal set by the model-and-data scale; the reward shape
nonetheless separates the runs on held-out EM. The
dominant residual failure mode in the rollouts we inspected is failure to
resolve the \emph{bridge entity} linking one hop to the next; no shape fixes it,
and we did not quantify its share.

\section{Limitations and Future Work}\label{sec:limitations}

Our results rest on a \textbf{single model family and size}: every trained
checkpoint is Qwen3.5-0.8B, with no cross-architecture or cross-scale
validation, so the findings are about one small model, not retrieval-augmented
RL in general. Training was likewise confined to a \textbf{single dataset}, a
precedented choice (ReSearch \cite{chen2025researchlearningreason}
also trains on MuSiQue alone), but we do not ablate data diversity, so
the reward-shape effect is not separated from the training distribution. The
\textbf{training horizon} is limited: the matched horizon is 180 steps, about
0.58 of one epoch, and several runs were still improving when stopped. That cuts
against our headline as well as for it, since sparse rewards can be slow
starters and seed-44 EM-only's late recovery to 0.318 fits a slow-start reading
we cannot exclude. Our \textbf{seed count is small}: three seeds suffice for the
exact test on the headline last-place ranking ($p \approx 0.037$; conservative
variants 0.074 and 0.111) but not for tight variance estimates, so the seed-44
inversion between the two F1 shapes is left open rather than resolved.
\textbf{Checkpoint selection is not independent of reporting}: each cell is the
best of roughly 18 to 31 per-checkpoint evaluations chosen on the same suite it
is reported on, with no held-out validation split. Since the per-checkpoint
series are noisy this inflates every cell, potentially unevenly across arms, and
of the per-seed EM-only gaps only seed-44's 6.9 points is clearly larger than
that effect. A selection rule independent of the reported score would most
strengthen the result. \textbf{Several recipe settings
are untested confounds for the mechanism}: the probability that a group
degenerates to zero reward variance is itself a function of the group size, so a
larger group than our $G=5$ may be the operative fix rather than a denser
reward. We ran \textbf{no supervised control}, so C1 is about this pipeline rather
than RL specifically, and we report only \textbf{automatic surface metrics},
with no human or judge-based evaluation of reasoning-chain quality. Finally, all three shapes plateau at the same \textbf{training-time exact-match
rate of 20 to 22\%} on MuSiQue rollouts, which we attribute to the
model-and-data scale rather than the reward (Section~\ref{sec:ceiling}); it
bounds the training signal, not the held-out averages.

These point to a coherent set of follow-ups: a \emph{same-generation control},
separating what Appendix~\ref{app:crossgen} owes to model generation versus
parameter count; a \emph{2B-or-above run}, testing whether ``EM-only is worst''
persists or inverts with scale; a \emph{semantic-similarity reward}
\cite{reimers2019sbert,zhang2020bertscore}; and retrieval-side levers on the
ceiling.

\section{Conclusion}\label{sec:conclusion}

We asked whether RL with verifiable rewards over a search tool can be made to
work on a sub-1B language model, and which reward shape is best. The feasibility
answer is yes: GRPO-training Qwen3.5-0.8B with an interleaved search tool for up
to one epoch on a single multi-hop corpus, on one four-GPU node, raises held-out
average exact match from a 0.092 untrained floor to between 0.271 and 0.307 on
average at the matched horizon, and to 0.352 in the best run, across a seven-benchmark
suite of which six were never trained on, under plain GRPO with no distillation
step.

The reward-shape answer is a negative result about the field's default. With the
rest of the recipe held fixed, the Search-R1-faithful exact-match-only reward is
the worst of the three at every seed at the matched horizon, worst even on the
exact-match metric it directly optimises, and the least seed-stable there.
Between the two partial-credit shapes the ordering is seed-dependent, so we
report a trade-off rather than a ranking: F1+format for seed-robustness, F1-only
for the ceiling. We propose a mechanism for the deficit, resting on the
within-group reward spread GRPO learns from, but do not establish it
(Section~\ref{sec:mechanism}). Bounded by one model, one corpus, and three seeds
(Section~\ref{sec:limitations}), the practical reading is that small-model RLVR
needs its own reward-design study rather than a scaled-down copy of a
large-model recipe.

\begin{credits}
\subsubsection{\ackname}
Training ran on a four-GPU RTX PRO 6000 Blackwell node provided by a CloudRift
compute grant. We thank the maintainers of NeMo-RL \cite{nemo_rl} and FlashRAG
\cite{flashrag}.

\subsubsection{\discintname}
The authors have no competing interests to declare.
\end{credits}

\FloatBarrier
\bibliographystyle{splncs04}
\bibliography{references}

\begin{thebibliography}{10}
\providecommand{\url}[1]{\texttt{#1}}
\providecommand{\urlprefix}{URL }
\providecommand{\doi}[1]{https://doi.org/#1}

\bibitem{asai2024selfrag}
Asai, A., Wu, Z., Wang, Y., Sil, A., Hajishirzi, H.: Self-{RAG}: Learning to
  retrieve, generate, and critique through self-reflection. In: International
  Conference on Learning Representations (ICLR) (2024), arXiv:2310.11511 (Oral)

\bibitem{belcak2025smalllanguagemodels}
Belcak, P., Heinrich, G., Diao, S., Fu, Y., Dong, X., Muralidharan, S., Lin,
  Y.C., Molchanov, P.: Small language models are the future of agentic ai
  (2025), \url{https://arxiv.org/abs/2506.02153}

\bibitem{chen2025researchlearningreason}
Chen, M., Sun, L., Li, T., Sun, H., Zhou, Y., Zhu, C., Wang, H., Pan, J.Z.,
  Zhang, W., Chen, H., Yang, F., Zhou, Z., Chen, W.: Research: Learning to
  reason with search for {LLMs} via reinforcement learning. In: Advances in
  Neural Information Processing Systems (NeurIPS) (2025), arXiv:2503.19470

\bibitem{Guo_2025}
{DeepSeek-AI}, Guo, D., Yang, D., Zhang, H., Song, J., Wang, P., Zhu, Q., Xu,
  R., et~al.: Deepseek-r1: Incentivizing reasoning capability in {LLMs} via
  reinforcement learning. Nature  \textbf{645},  633--638 (2025).
  \doi{10.1038/s41586-025-09422-z}, arXiv:2501.12948

\bibitem{douze2024faiss}
Douze, M., Guzhva, A., Deng, C., Johnson, J., Szilvasy, G., Mazar{\'e}, P.E.,
  Lomeli, M., Hosseini, L., J{\'e}gou, H.: The {Faiss} library (2024),
  \url{https://arxiv.org/abs/2401.08281}

\bibitem{fatemi2025concise}
Fatemi, M., Rafiee, B., Tang, M., Talamadupula, K.: Concise reasoning via
  reinforcement learning (2025), \url{https://arxiv.org/abs/2504.05185}

\bibitem{justrl2025}
He, Qu, Liu, et~al.: {JustRL}: Scaling a 1.5{B} {LLM} with a simple {RL}
  recipe. ICLR 2026 Blogpost Track (2025),
  \url{https://arxiv.org/abs/2512.16649}, arXiv:2512.16649

\bibitem{2WikiMultiHopQA}
Ho, X., Duong~Nguyen, A.K., Sugawara, S., Aizawa, A.: Constructing a multi-hop
  {QA} dataset for comprehensive evaluation of reasoning steps. In: Proceedings
  of the 28th International Conference on Computational Linguistics (COLING)
  (2020), arXiv:2011.01060 (2WikiMultiHopQA)

\bibitem{jiang2023flare}
Jiang, Z., Xu, F.F., Gao, L., Sun, Z., Liu, Q., Dwivedi-Yu, J., Yang, Y.,
  Callan, J., Neubig, G.: Active retrieval augmented generation. In:
  Proceedings of the 2023 Conference on Empirical Methods in Natural Language
  Processing (EMNLP) (2023), arXiv:2305.06983 (FLARE)

\bibitem{Jin2025SearchR1TL}
Jin, B., Zeng, H., Yue, Z., Yoon, J., Arik, S., Wang, D., Zamani, H., Han, J.:
  Search-r1: Training {LLMs} to reason and leverage search engines with
  reinforcement learning. In: Conference on Language Modeling (COLM) (2025),
  arXiv:2503.09516

\bibitem{flashrag}
Jin, J., Zhu, Y., Yang, X., Zhang, C., Dou, Z.: {FlashRAG}: A modular toolkit
  for efficient retrieval-augmented generation research. In: Companion
  Proceedings of the ACM Web Conference (WWW), Resource Track (2025).
  \doi{10.1145/3701716.3715313}, arXiv:2405.13576

\bibitem{TriviaQA}
Joshi, M., Choi, E., Weld, D.S., Zettlemoyer, L.: {TriviaQA}: A large scale
  distantly supervised challenge dataset for reading comprehension. In:
  Proceedings of the 55th Annual Meeting of the Association for Computational
  Linguistics (ACL) (2017), arXiv:1705.03551

\bibitem{karpukhin2020dpr}
Karpukhin, V., O{\u{g}}uz, B., Min, S., Lewis, P., Wu, L., Edunov, S., Chen,
  D., Yih, W.t.: Dense passage retrieval for open-domain question answering.
  In: Proceedings of the 2020 Conference on Empirical Methods in Natural
  Language Processing (EMNLP). pp. 6769--6781 (2020),
  \url{https://aclanthology.org/2020.emnlp-main.550/}

\bibitem{kotoge2025dgpo}
Kotoge, R., Nishimura, M., Ma, J.: Can compact language models search like
  agents? distillation-guided policy optimization for preserving agentic {RAG}
  capabilities. In: Proceedings of the Annual Meeting of the Association for
  Computational Linguistics (ACL) (2026), arXiv:2508.20324 (DGPO)

\bibitem{natural_questions_nq}
Kwiatkowski, T., Palomaki, J., Redfield, O., Collins, M., Parikh, A., Alberti,
  C., Epstein, D., Polosukhin, I., Devlin, J., Lee, K., et~al.: Natural
  questions: A benchmark for question answering research. Transactions of the
  Association for Computational Linguistics (TACL)  \textbf{7},  452--466
  (2019)

\bibitem{lambert2024tulu3}
Lambert, N., Morrison, J., Pyatkin, V., Huang, S., Ivison, H., Brahman, F.,
  et~al.: {Tulu} 3: Pushing frontiers in open language model post-training. In:
  Conference on Language Modeling (COLM) (2025), arXiv:2411.15124

\bibitem{Lewis2020rag}
Lewis, P., Perez, E., Piktus, A., Petroni, F., Karpukhin, V., Goyal, N.,
  K{\"u}ttler, H., Lewis, M., Yih, W.t., Rockt{\"a}schel, T., Riedel, S.,
  Kiela, D.: Retrieval-augmented generation for knowledge-intensive {NLP}
  tasks. In: Advances in Neural Information Processing Systems (NeurIPS)
  (2020), arXiv:2005.11401

\bibitem{rorecomp2025}
Li, G., Qin, Y., Tan, X., Yang, D., Shi, Y., Xu, Z., Li, X., Sun, X., Li, K.:
  {RoRecomp}: Enhancing reasoning efficiency via rollout response recomposition
  in reinforcement learning (2025), \url{https://arxiv.org/abs/2509.25958},
  arXiv:2509.25958

\bibitem{IG-Search}
Liang, Z., Ma, Y., Chen, B., Qian, Z., Dai, H., Mao, L., Zhang, X., Lei, C.,
  Ou, W.: {IG-Search}: Step-level information gain rewards for search-augmented
  reasoning (2026), \url{https://arxiv.org/abs/2604.15148}, arXiv:2604.15148

\bibitem{liu2025drgrpo}
Liu, Z., Chen, C., Li, W., Pang, T., Du, C., Lin, M.: Understanding
  r1-zero-like training: A critical perspective. In: Conference on Language
  Modeling (COLM) (2025), arXiv:2503.20783 (Dr. GRPO)

\bibitem{PopQA}
Mallen, A., Asai, A., Zhong, V., Das, R., Khashabi, D., Hajishirzi, H.: When
  not to trust language models: Investigating effectiveness of parametric and
  non-parametric memories. In: Proceedings of the 61st Annual Meeting of the
  Association for Computational Linguistics (ACL) (2023), arXiv:2212.10511
  (PopQA)

\bibitem{nemo_rl}
{NVIDIA NeMo Team}: {NeMo RL}: A scalable and efficient post-training library.
  \url{https://github.com/NVIDIA-NeMo/RL} (2025)

\bibitem{openai2024learning}
{OpenAI}: Learning to reason with {LLMs}.
  \url{https://openai.com/index/learning-to-reason-with-llms/} (2024),
  accessed: 2026-06-09

\bibitem{ouyang2022instructgpt}
Ouyang, L., Wu, J., Jiang, X., Almeida, D., Wainwright, C.L., Mishkin, P.,
  Zhang, C., Agarwal, S., Slama, K., Ray, A., et~al.: Training language models
  to follow instructions with human feedback. In: Advances in Neural
  Information Processing Systems (NeurIPS) (2022), arXiv:2203.02155

\bibitem{press2023compositionality}
Press, O., Zhang, M., Min, S., Schmidt, L., Smith, N.A., Lewis, M.: Measuring
  and narrowing the compositionality gap in language models. In: Findings of
  the Association for Computational Linguistics: EMNLP 2023 (2023),
  arXiv:2210.03350 (Self-Ask; Bamboogle)

\bibitem{qwen3.5}
{Qwen Team}: Qwen3.5: Towards native multimodal agents. Alibaba Qwen blog and
  Hugging Face model cards (Qwen/Qwen3.5-0.8B) (2026),
  \url{https://qwen.ai/blog?id=qwen3.5}

\bibitem{radford2019gpt2}
Radford, A., Wu, J., Child, R., Luan, D., Amodei, D., Sutskever, I.: Language
  models are unsupervised multitask learners. OpenAI technical report (2019),
  \url{https://cdn.openai.com/better-language-models/language_models_are_unsupervised_multitask_learners.pdf}

\bibitem{rajpurkar2016squad}
Rajpurkar, P., Zhang, J., Lopyrev, K., Liang, P.: {SQuAD}: 100,000+ questions
  for machine comprehension of text. In: Proceedings of the 2016 Conference on
  Empirical Methods in Natural Language Processing (EMNLP). pp. 2383--2392
  (2016). \doi{10.18653/v1/D16-1264}, \url{https://aclanthology.org/D16-1264/}

\bibitem{reimers2019sbert}
Reimers, N., Gurevych, I.: Sentence-{BERT}: Sentence embeddings using siamese
  {BERT}-networks. In: Proceedings of the 2019 Conference on Empirical Methods
  in Natural Language Processing and the 9th International Joint Conference on
  Natural Language Processing (EMNLP-IJCNLP) (2019), arXiv:1908.10084

\bibitem{toolformer}
Schick, T., Dwivedi-Yu, J., Dess{\`i}, R., Raileanu, R., Lomeli, M., Hambro,
  E., Zettlemoyer, L., Cancedda, N., Scialom, T.: Toolformer: Language models
  can teach themselves to use tools. In: Advances in Neural Information
  Processing Systems (NeurIPS) (2023), arXiv:2302.04761

\bibitem{schulman2017ppo}
Schulman, J., Wolski, F., Dhariwal, P., Radford, A., Klimov, O.: Proximal
  policy optimization algorithms (2017), \url{https://arxiv.org/abs/1707.06347}

\bibitem{shao2024deepseekmath}
Shao, Z., Wang, P., Zhu, Q., Xu, R., Song, J., Zhang, M., Li, Y.K., Wu, Y.,
  Guo, D.: Deepseekmath: Pushing the limits of mathematical reasoning in open
  language models (2024), \url{https://arxiv.org/abs/2402.03300}

\bibitem{sheng2024hybridflow}
Sheng, G., Zhang, C., Ye, Z., Wu, X., Zhang, W., Zhang, R., Peng, Y., Lin, H.,
  Wu, C.: Hybridflow: A flexible and efficient {RLHF} framework (2024).
  \doi{10.1145/3689031.3696075}, \url{https://arxiv.org/abs/2409.19256}, verl;
  published at EuroSys 2025

\bibitem{song2025r1searcher}
Song, H., Jiang, J., Min, Y., Chen, J., Chen, Z., Zhao, W.X., Fang, L., Wen,
  J.R.: R1-searcher: Incentivizing the search capability in {LLMs} via
  reinforcement learning (2025), \url{https://arxiv.org/abs/2503.05592}

\bibitem{MuSiQue}
Trivedi, H., Balasubramanian, N., Khot, T., Sabharwal, A.: {MuSiQue}: Multihop
  questions via single-hop question composition. Transactions of the
  Association for Computational Linguistics (TACL)  \textbf{10},  539--554
  (2022), arXiv:2108.00573

\bibitem{ircot}
Trivedi, H., Balasubramanian, N., Khot, T., Sabharwal, A.: Interleaving
  retrieval with chain-of-thought reasoning for knowledge-intensive multi-step
  questions. In: Proceedings of the 61st Annual Meeting of the Association for
  Computational Linguistics (ACL) (2023), arXiv:2212.10509 (IRCoT)

\bibitem{wang2025tina}
Wang, Asilis, Akg{\"u}l, et~al.: Tina: Tiny reasoning models via {LoRA} (2025),
  \url{https://arxiv.org/abs/2504.15777}

\bibitem{wang2025oneshotrlvr}
Wang, Yang, Zeng, et~al.: Reinforcement learning for reasoning in large
  language models with one training example (2025),
  \url{https://arxiv.org/abs/2504.20571}

\bibitem{wang2024comprehensivesurveysmalllanguage}
Wang, F., Zhang, Z., Zhang, X., Wu, Z., Mo, T., Lu, Q., Wang, W., Li, R., Xu,
  J., Tang, X., He, Q., Ma, Y., Huang, M., Wang, S.: A comprehensive survey of
  small language models in the era of large language models: Techniques,
  enhancements, applications, collaboration with llms, and trustworthiness
  (2024), \url{https://arxiv.org/abs/2411.03350}

\bibitem{IGPO}
Wang, G., Dai, S., Ye, G., Gan, Z., Yao, W., Deng, Y., Wu, X., Ying, Z.:
  Information gain-based policy optimization: A simple and effective approach
  for multi-turn search agents. In: International Conference on Learning
  Representations (ICLR) (2026), arXiv:2510.14967 (IGPO)

\bibitem{otc2025}
Wang, H., Qian, C., Zhong, W., Chen, X., Qiu, J., Huang, S., Jin, B., Wang, M.,
  Wong, K.F., Ji, H.: Acting less is reasoning more! teaching model to act
  efficiently (2025), \url{https://arxiv.org/abs/2504.14870}, arXiv:2504.14870
  (OTC / OTC-PO)

\bibitem{wang2022e5}
Wang, L., Yang, N., Huang, X., Jiao, B., Yang, L., Jiang, D., Majumder, R.,
  Wei, F.: Text embeddings by weakly-supervised contrastive pre-training
  (2022), \url{https://arxiv.org/abs/2212.03533}, e5

\bibitem{wei2022cot}
Wei, J., Wang, X., Schuurmans, D., Bosma, M., Ichter, B., Xia, F., Chi, E.H.,
  Le, Q.V., Zhou, D.: Chain-of-thought prompting elicits reasoning in large
  language models. In: Advances in Neural Information Processing Systems
  (NeurIPS) (2022), arXiv:2201.11903

\bibitem{wu2025hiprag}
Wu, P., Zhang, M., Wan, K., Zhao, W., He, K., Du, X., Chen, Z.: {HiPRAG}:
  Hierarchical process rewards for efficient agentic retrieval augmented
  generation. In: International Conference on Learning Representations (ICLR)
  (2026), arXiv:2510.07794

\bibitem{Search-P1}
Xia, T., Xu, M., Hu, L., Sun, Y., Li, W., Shang, L., Liu, L., Shu, P., Yu, H.,
  Jiang, J.: {Search-P1}: Path-centric reward shaping for stable and efficient
  agentic {RAG} training. In: Proceedings of the Association for Computational
  Linguistics (ACL), Industry Track (2026), arXiv:2602.22576

\bibitem{TIPS}
Xie, Y., Thomas, N., Hansen, N., Fu, Y., Li, L.E., Wang, X.: {TIPS}: Turn-level
  information-potential reward shaping for search-augmented {LLMs}. In:
  International Conference on Learning Representations (ICLR) (2026),
  arXiv:2603.22293

\bibitem{xu2026howtotrain}
Xu, Y., Lu, S., Cheng, J., Wang, M., Xie, Q., Wang, X., He, R., Liang, J.: How
  to train your deep research agent? prompt, reward, and policy optimization in
  {Search-R1} (2026), \url{https://arxiv.org/abs/2602.19526}, arXiv:2602.19526

\bibitem{HotpotQA}
Yang, Z., Qi, P., Zhang, S., Bengio, Y., Cohen, W.W., Salakhutdinov, R.,
  Manning, C.D.: {HotpotQA}: A dataset for diverse, explainable multi-hop
  question answering. In: Proceedings of the 2018 Conference on Empirical
  Methods in Natural Language Processing (EMNLP) (2018), arXiv:1809.09600

\bibitem{react_yao_2023}
Yao, S., Zhao, J., Yu, D., Du, N., Shafran, I., Narasimhan, K., Cao, Y.:
  {ReAct}: Synergizing reasoning and acting in language models. In:
  International Conference on Learning Representations (ICLR) (2023),
  arXiv:2210.03629

\bibitem{yu2025dapo}
Yu, Q., Zhang, Z., Zhu, R., Yuan, Y., Zuo, X., Yue, Y., et~al.: {DAPO}: An
  open-source {LLM} reinforcement learning system at scale (2025),
  \url{https://arxiv.org/abs/2503.14476}

\bibitem{zhan2025exgrpo}
Zhan, Li, Wang, et~al.: {ExGRPO}: Learning to reason from experience. In:
  International Conference on Learning Representations (ICLR) (2026),
  arXiv:2510.02245

\bibitem{zhang2020bertscore}
Zhang, T., Kishore, V., Wu, F., Weinberger, K.Q., Artzi, Y.: {BERTScore}:
  Evaluating text generation with {BERT}. In: International Conference on
  Learning Representations (ICLR) (2020), arXiv:1904.09675

\bibitem{zheng2025gspo}
Zheng, C., Liu, S., Li, M., Chen, X.H., Yu, B., Lin, J., et~al.: Group sequence
  policy optimization (2025), \url{https://arxiv.org/abs/2507.18071}

\end{thebibliography}

\appendix
\renewcommand{\theHsection}{app.\thesection}

\section{Per-seed Held-out Trajectories}\label{app:perseed}

\begin{figure}[htbp]
\centering
\includegraphics[width=0.86\linewidth]{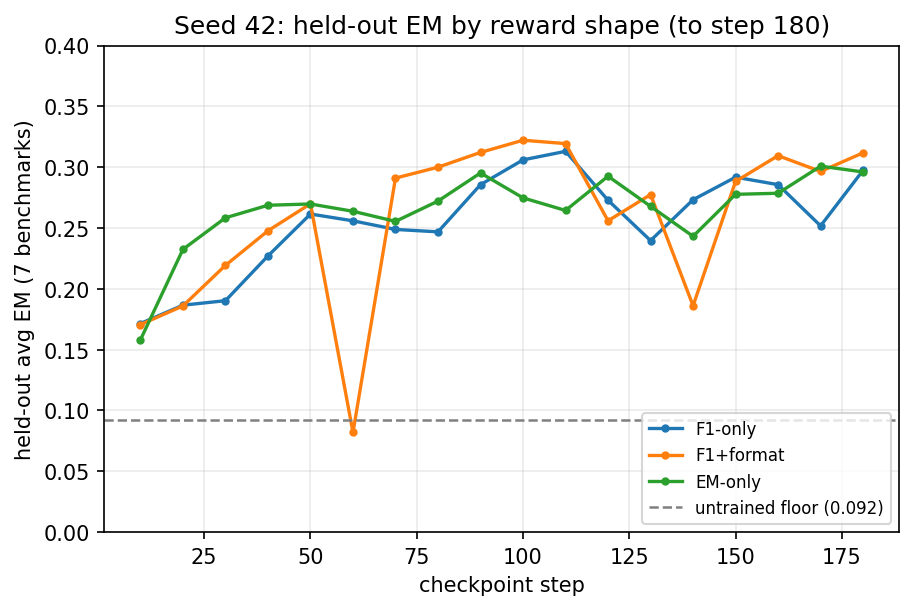}
\caption{Seed 42 held-out EM by reward shape (to step 180); F1+format leads and
all three shapes clear the 0.092 floor early.}
\label{fig:app-s42}
\end{figure}

\begin{figure}[htbp]
\centering
\includegraphics[width=0.86\linewidth]{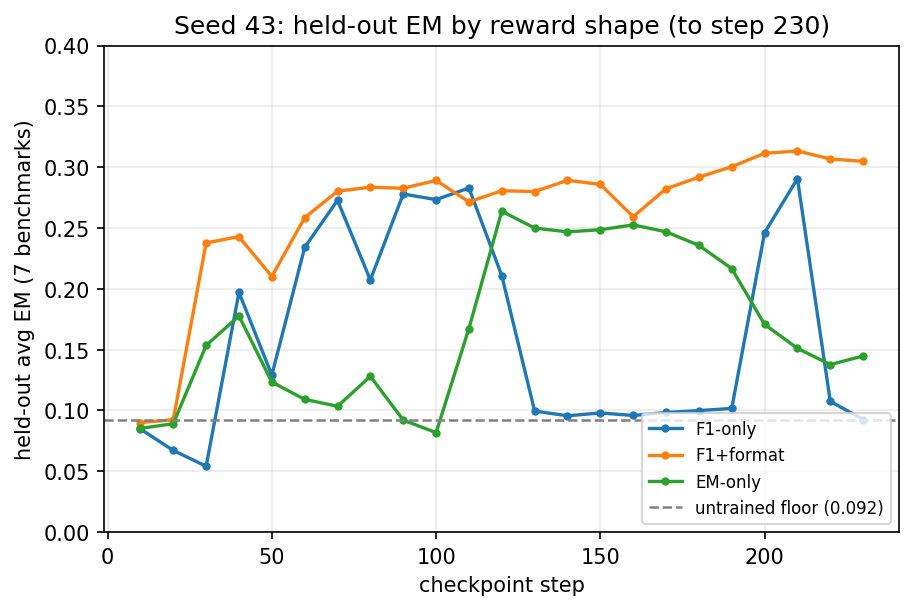}
\caption{Seed 43 held-out EM by reward shape (to step 230); F1+format leads and
is the most stable, while F1-only is the noisiest leg, with a deep collapse over
steps 130 to 190.}
\label{fig:app-s43}
\end{figure}

\begin{figure}[htbp]
\centering
\includegraphics[width=0.86\linewidth]{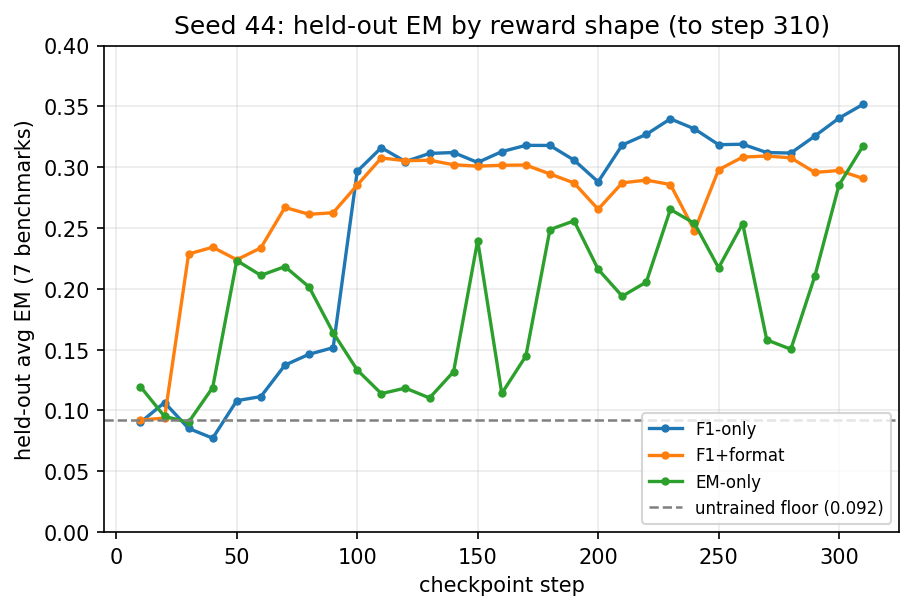}
\caption{Seed 44 held-out EM by reward shape (full epoch, to step 310); F1-only
reaches the run-high 0.352 at step 310, and EM-only rises late, on the longest
leg in the study.}
\label{fig:app-s44}
\end{figure}

Figures~\ref{fig:app-s42} to~\ref{fig:app-s44} show one seed each, overlaying
the three reward shapes against the untrained floor. Seed 44 runs the
full epoch and is where EM-only's late rise appears.

\FloatBarrier
\section{Training Dynamics}\label{app:dyn}

\begin{figure}[htbp]
\centering
\includegraphics[width=0.88\linewidth]{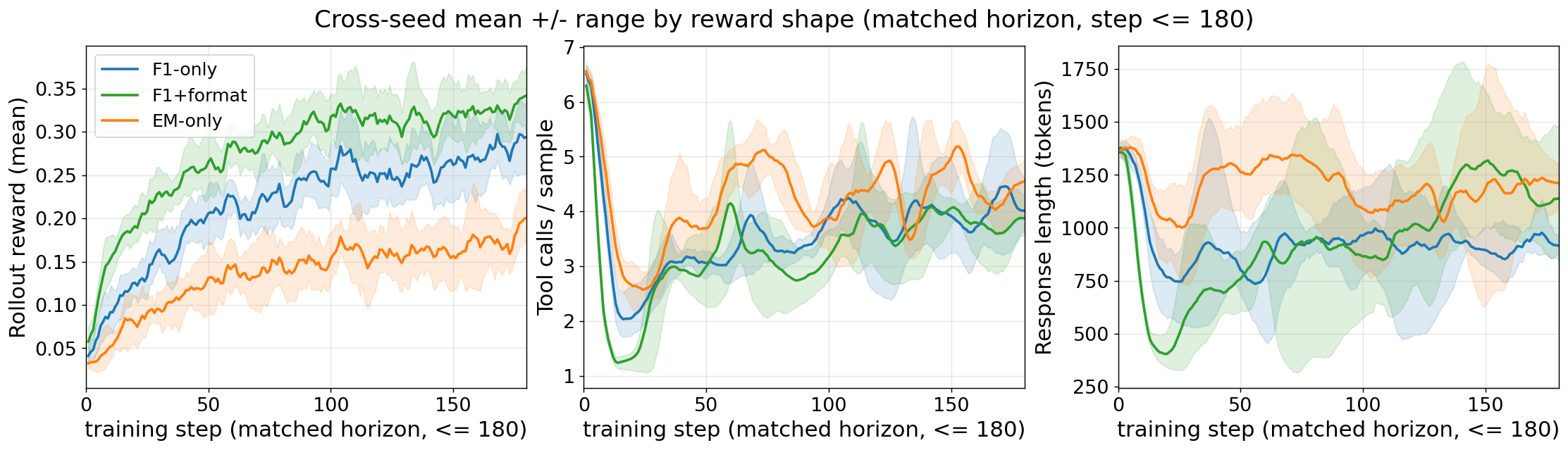}
\caption{Cross-seed mean with minimum-to-maximum range per reward shape, matched
to step $\le 180$, for reward, tool calls, and response length. The F1 shapes
lean while reward climbs; EM-only stays highest on tool calls (reward levels
are not comparable across shapes, Section~\ref{sec:ceiling}).}
\label{fig:dyn}
\end{figure}

\begin{figure}[htbp]
\centering
\includegraphics[width=0.92\linewidth]{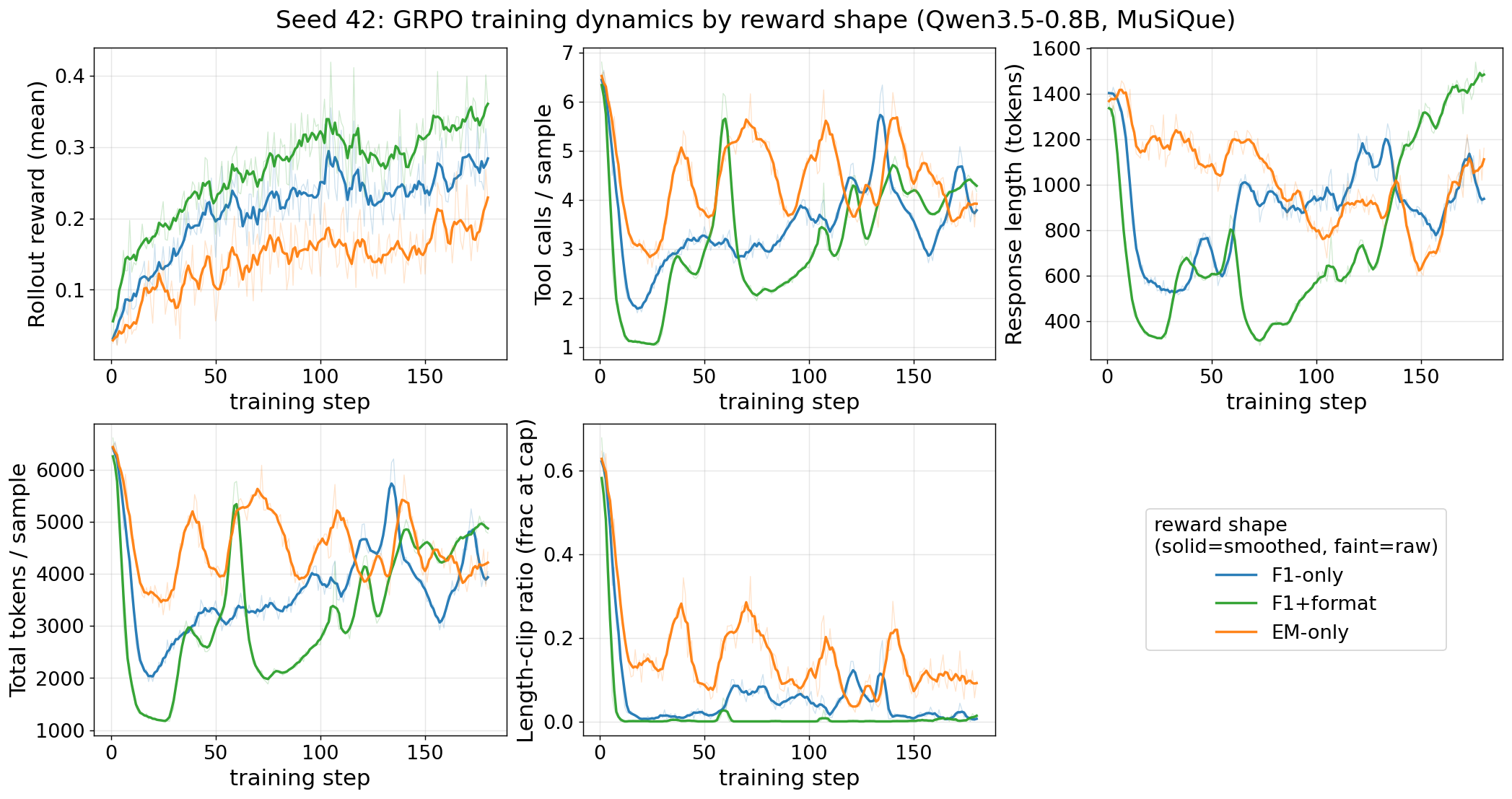}
\caption{Seed 42 training dynamics, capped at step 180.}
\label{fig:app-dyn42}
\end{figure}

\begin{figure}[htbp]
\centering
\includegraphics[width=0.92\linewidth]{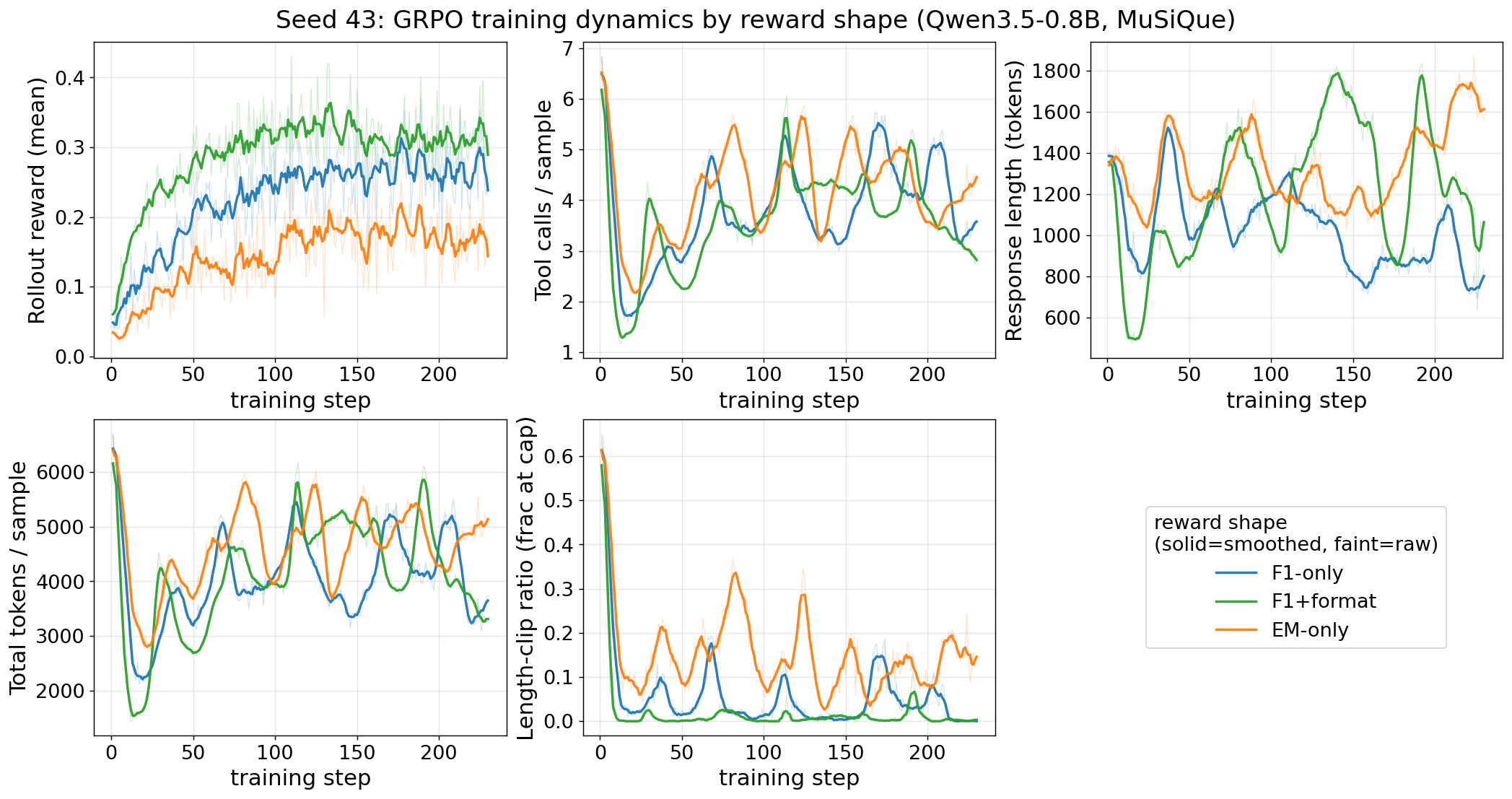}
\caption{Seed 43 training dynamics, capped at step 230.}
\label{fig:app-dyn43}
\end{figure}

\begin{figure}[htbp]
\centering
\includegraphics[width=0.92\linewidth]{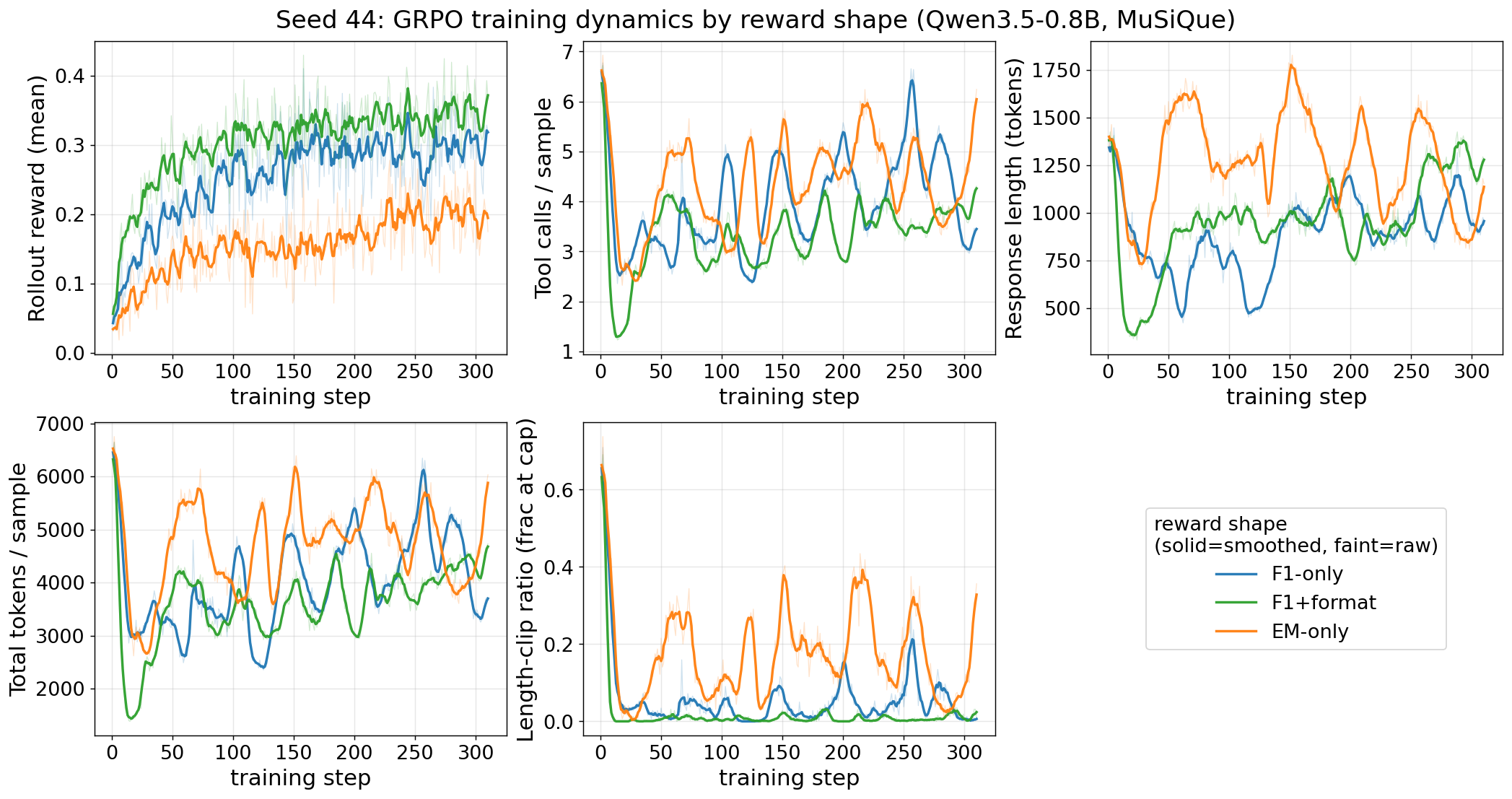}
\caption{Seed 44 training dynamics (full epoch; last full cadence at step 310).
Reward climbs while tool calls, length, total tokens, and clip ratio all fall
for the F1 shapes, while EM-only stays high.}
\label{fig:app-dyn44}
\end{figure}

Figure~\ref{fig:dyn} is the cross-seed summary discussed in
Section~\ref{sec:mechanism}. Figures~\ref{fig:app-dyn42}
to~\ref{fig:app-dyn44} then overlay the three reward shapes on reward, tool
calls per sample, response length, total tokens per sample, and length-clip
ratio, one seed per figure. Each line is a centred moving average drawn over the
faint raw per-step series, so both the trend and the background noise are
visible.

\FloatBarrier
\section{Emergent-compression Detail}\label{app:compression}

Table~\ref{tab:app-compression} gives the cold-start-to-end magnitudes for the
full-epoch seed-44 leg (step 1 to step 310) behind
Section~\ref{sec:compression}. The no-valid-answer (aborted) rate tracks the
length-clip ratio almost exactly in every run (Pearson $+0.98$ to $+0.99$), so
hitting the generation cap and failing to produce a scorable answer are
effectively the same event; at cold start 62 to 68\% of rollouts hit the cap and
72 to 75\% emit no valid answer.

Reward anti-correlates with the length-clip ratio, computed over each leg
truncated to its seed cap as everywhere else in this paper, at Pearson $-0.50$,
$-0.45$, $-0.44$ (F1-only) and $-0.47$, $-0.52$, $-0.49$ (F1+format) at seeds
42, 43 and 44. For EM-only the corresponding values are $-0.50$, $-0.38$ and
$-0.20$. The weakening that the cap-as-implicit-penalty reading predicts is
therefore clear only at seed 44, the one leg on which EM-only does not stay
compressed; at seed 42 EM-only sits inside the F1 range. We report the third
measurement of Section~\ref{sec:compression} as seed-44 evidence rather than a
cross-seed regularity. (The seed-42 EM-only leg is the one case where the
truncation matters: read to its full 211 steps instead of its 180-step cap it
falls to $-0.33$, outside the F1 range.)

\begin{table}[htbp]
\centering
\caption{Seed-44 cold-start-to-end magnitudes (step 1 to step 310). The F1 shapes
compress on every axis shown; EM-only leans mid-horizon and then re-enters over-search. Per-rollout response length is not shown because it does not move in one
direction across shapes (Section~\ref{sec:compression}).}
\label{tab:app-compression}
\begin{tabular}{lrrr}
\toprule
Quantity (step 1 $\rightarrow$ 310) & F1-only & F1+format & EM-only \\
\midrule
Tool calls per sample   & 6.6 $\rightarrow$ 3.4 & 6.5 $\rightarrow$ 4.3 & 6.5 $\rightarrow$ 6.2 \\

Total tokens per sample & 6435 $\rightarrow$ 3694 & 6360 $\rightarrow$ 4730 & 6380 $\rightarrow$ 6031 \\
Length-clip ratio       & 0.63 $\rightarrow$ 0.01 & 0.63 $\rightarrow$ 0.01 & 0.62 $\rightarrow$ 0.36 \\
No-valid-answer ratio   & 0.72 $\rightarrow$ 0.01 & 0.72 $\rightarrow$ 0.01 & 0.72 $\rightarrow$ 0.37 \\
\bottomrule
\end{tabular}
\end{table}

The residual share of rollouts still at the generation cap at the last full
cadence separates the shapes cleanly: EM-only leaves 9.7\%, 14.1\%, and 35.6\%
at seeds 42, 43, and 44 respectively, against about 1\% for both F1 shapes.

\FloatBarrier
\section{Held-out Bars and Metric Robustness}\label{app:metric}

\begin{figure}[htbp]
\centering
\includegraphics[width=0.86\linewidth]{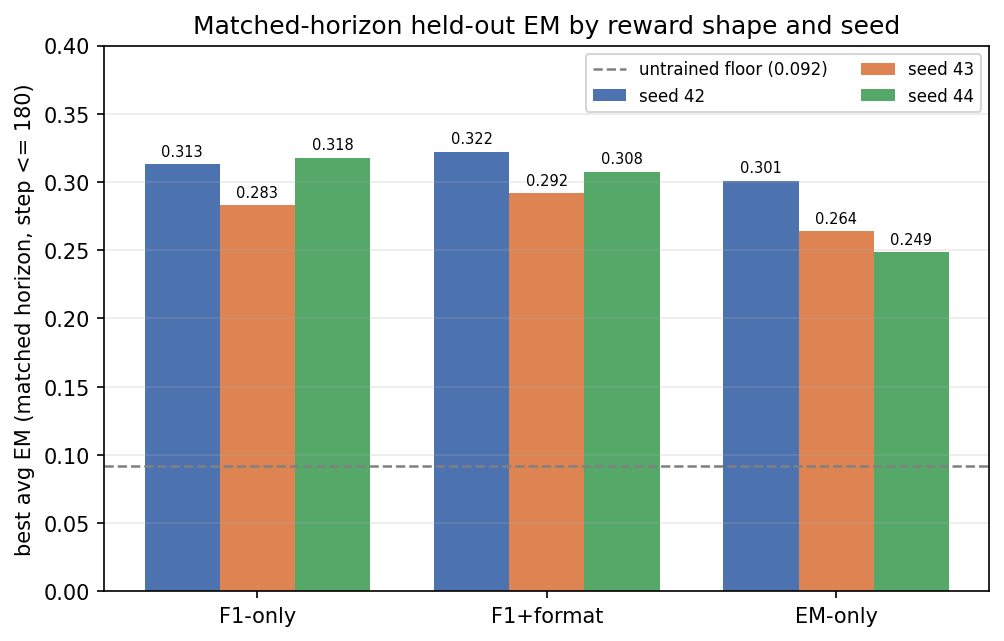}
\caption{Matched-horizon average EM by reward shape, one bar per seed (the same
cells as the upper panel of Table~\ref{tab:matched}). EM-only is the shortest group at every seed;
the bar spread shows seed variance (F1+format tightest, EM-only widest). All
cells are best checkpoints under the same step $\le 180$ cap, so this is not a
training-length effect.}
\label{fig:heldout-bar}
\end{figure}

\begin{figure}[htbp]
\centering
\includegraphics[width=0.86\linewidth]{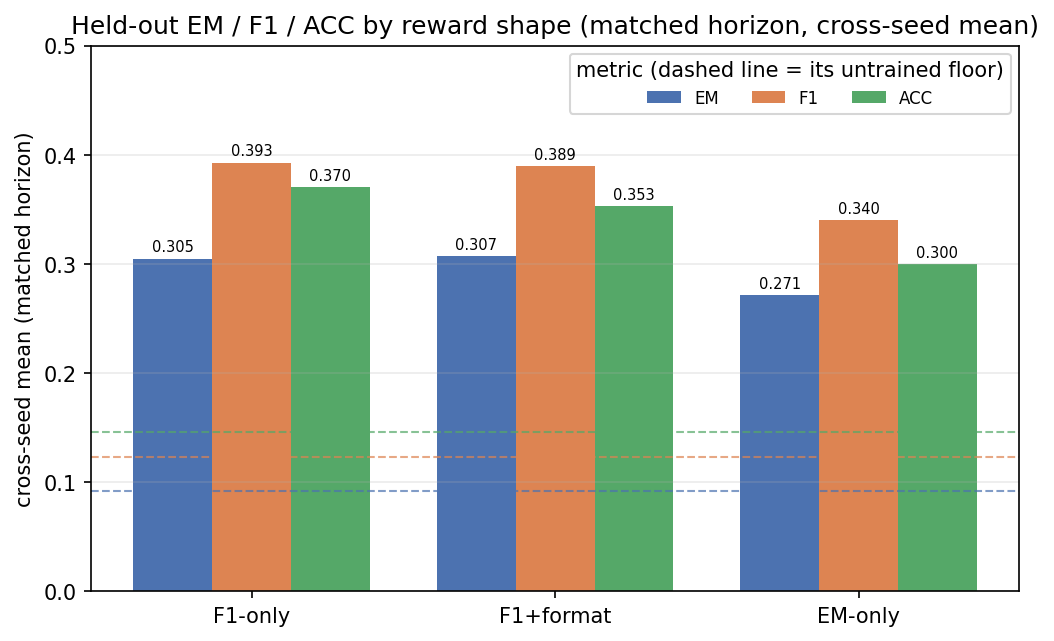}
\caption{Matched-horizon cross-seed mean of EM, token-F1, and substring accuracy
at each run's best-EM checkpoint, by reward shape; dashed lines mark each
metric's untrained floor. EM-only is lowest on all three metrics and the two F1
shapes are close and higher, so the ranking is not an artifact of scoring on the
metric EM-only optimises.}
\label{fig:3metric}
\end{figure}

Figure~\ref{fig:heldout-bar} plots the nine matched-horizon cells of
Table~\ref{tab:matched} as bars, and Figure~\ref{fig:3metric} repeats the
matched-horizon comparison on all three evaluation metrics.

\FloatBarrier
\section{Cross-generation Context}\label{app:crossgen}

Our Search-R1 reproduction runs (Section~\ref{sec:results}) also supply a
reference point one model generation back, which we record here as context. The
previous-generation Qwen2.5-3B (instruct) moves from a 0.199 untrained floor to
0.341 trained, against our 0.092 to 0.352, an endpoint the 0.8B model reaches on
roughly one twenty-sixth of the 3B checkpoint's training-prompt exposure. Read
narrowly, a current-generation sub-1B model plus RLVR matches, and nominally
exceeds, a roughly four-times-larger previous-generation model measured on the
same pipeline, and the far smaller training budget makes that reading more
conservative, not less. It is not a result. The two endpoints differ by 1.1
points, inside our own reproduction error, and the two settings differ in model,
generation, corpus, reward, and recipe, so nothing about the effect of RL as a
function of parameter count \emph{alone} follows.

\FloatBarrier
\section{Training-health Diagnostics}\label{app:health}

\begin{figure}[htbp]
\centering
\includegraphics[width=0.92\linewidth]{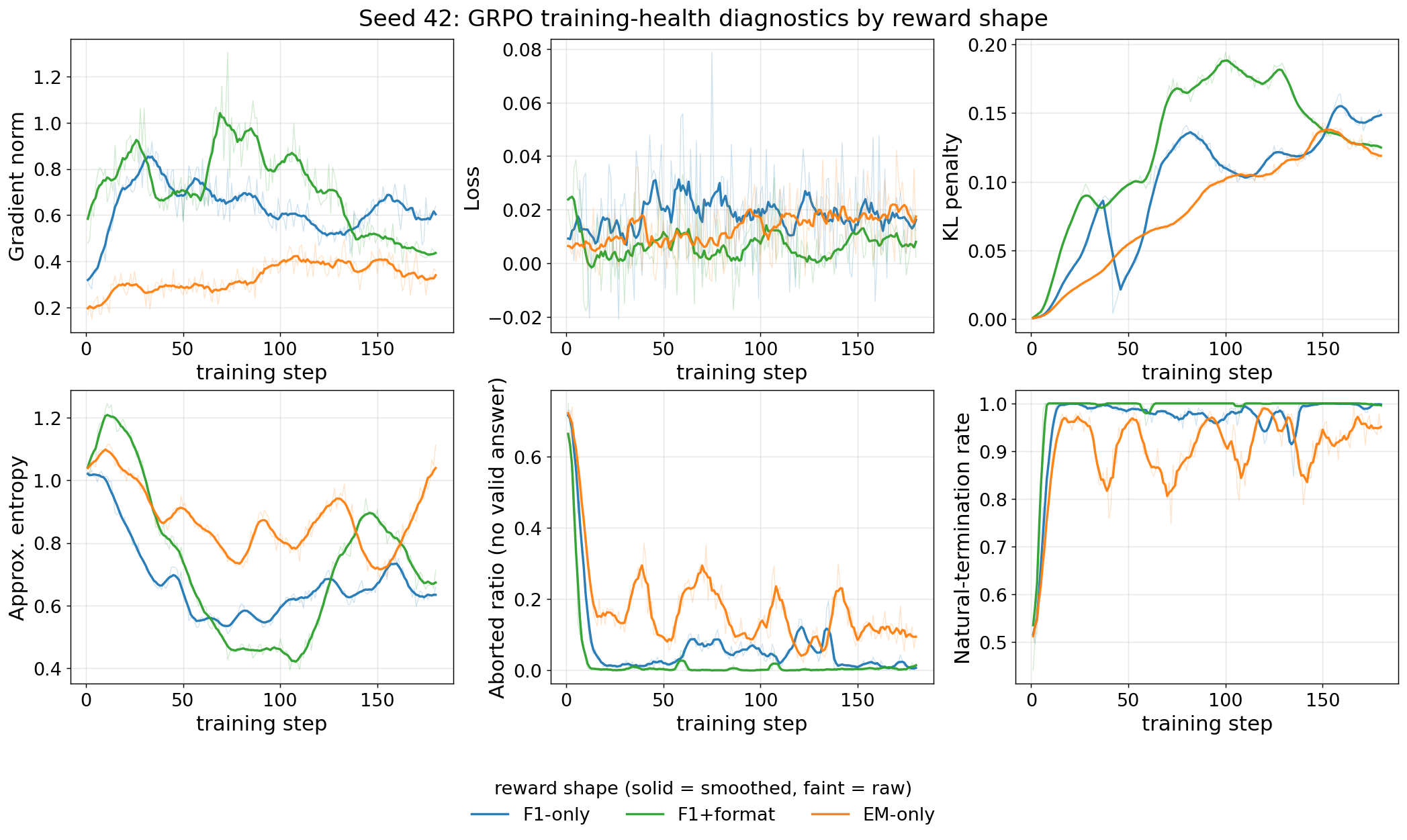}
\caption{Seed-42 training-health diagnostics by reward shape: gradient norm,
loss, KL divergence, approximate policy entropy, aborted ratio, and
natural-termination rate. Seed 42 is shown; the bounded-gradient-norm and
lowest-EM-only-gradient-norm reads are computed over all nine runs (the other
seeds, not shown, are similar).}
\label{fig:app-health}
\end{figure}

Figure~\ref{fig:app-health} plots the two quantities the mechanism in
Section~\ref{sec:mechanism} rests on, gradient norm and policy entropy, together
with loss, KL divergence, the aborted ratio, and the natural-termination rate,
for seed 42. Gradient norm stays bounded throughout; entropy declines
(sharpening) for the F1 shapes while EM-only stays pinned high. Note that the
entropy panel is an approximate token-level entropy logged during training, not
an exact policy entropy.

\FloatBarrier
\section{Worked Rollout and Loss Masking}\label{app:rollout}

\begin{figure}[htbp]
\centering
\includegraphics[trim={295 0 0 0},clip,width=0.70\linewidth]{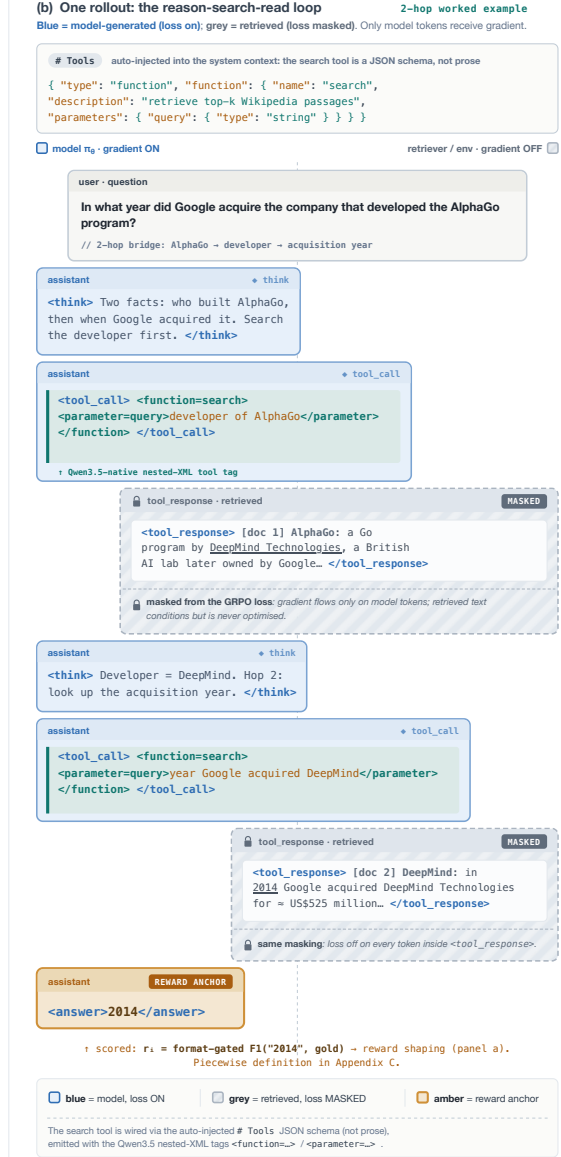}
\caption{One reason-search-read rollout, with the retrieved spans shaded to mark
the loss mask. The reward is computed on the final answer block only. This panel
and Figure~\ref{fig:pipeline} are two halves of one source drawing, so its
in-figure lettering is inherited: ``panel a'' is Figure~\ref{fig:pipeline}, and
the ``Appendix C'' it cites for the piecewise reward definition is a stale
pointer, superseded here by Equation~\ref{eq:rewards}.}
\label{fig:app-rollout}
\end{figure}

Figure~\ref{fig:app-rollout} shows one schematic reason-search-read rollout on a
two-hop question: the model interleaves reasoning with Qwen3.5-native tool calls,
reads retrieved passages, and ends in an answer. Retrieved spans are shaded to
mark that the GRPO loss is computed only on model-generated tokens. The
transcript is illustrative, not a verbatim MuSiQue trace.

\FloatBarrier
\section{Held-out Search-count Profile and the Termination Gap}
\label{app:termination}

\begin{figure}[htbp]
\centering
\includegraphics[width=\linewidth]{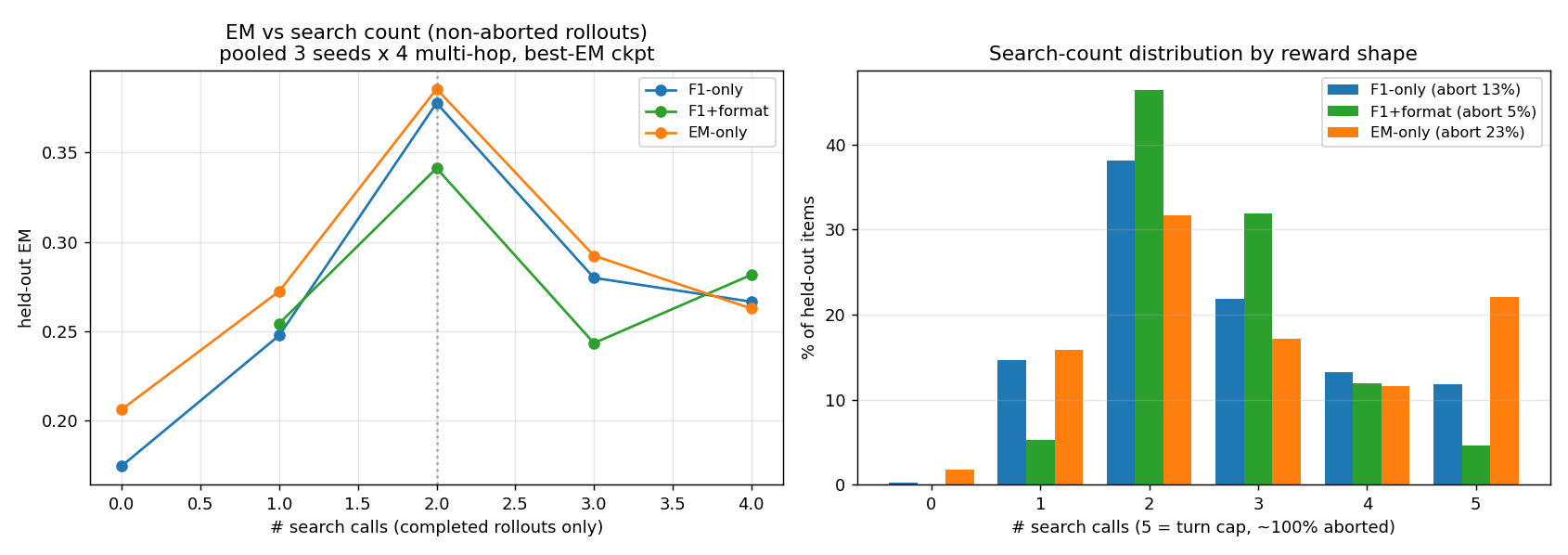}
\caption{Held-out behaviour by search count at each run's best-EM checkpoint,
pooled over three seeds and the four multi-hop benchmarks. Left: mean exact
match against search count for completed rollouts; all three shapes peak at two
searches. Right: the search-count distribution per shape with the per-shape
abort rate; EM-only places far more mass in the aborted turn-budget bin, the
main source of its lower held-out exact match.}
\label{fig:app-searchcount}
\end{figure}

At each run's best-EM checkpoint we recovered, for every held-out question on
the four multi-hop benchmarks (MuSiQue, 2WikiMultiHopQA, HotpotQA, Bamboogle), the
number of search calls the policy issued and whether the answer was an exact
match; pooling the three seeds gives 67{,}569 graded rollouts per reward shape
(Figure~\ref{fig:app-searchcount}). Held-out exact match against search count
is an inverted-U peaking at two searches for every shape and every dataset;
among completed rollouts, two searches scores 0.378 (F1-only), 0.341
(F1+format), and 0.386 (EM-only). The shapes separate by 1 to 5 points at a
given bin, but not in the direction that would explain the held-out gap: EM-only,
the weakest shape overall, is at or above both F1 shapes at the peak. Its
held-out deficit therefore does not come from answering worse at a given
completed search count. What differs is
termination: 22.8\% of EM-only rollouts abort (reach the turn budget or emit no
parseable answer) against 12.7\% for F1-only and 5.0\% for F1+format, and
EM-only places 22.0\% of its rollouts in the fully-aborted top search bin
against 4.6\% for F1+format. Three cautions apply. The steep drop at the top of
the curve is the truncation tail (those rollouts score zero by construction),
not over-searching degrading a finished answer. The milder decline from two to
four searches is confounded by question difficulty. And conditioning on
completion selects a different fraction of rollouts per shape. We therefore read
the curve as descriptive, not causal.

\FloatBarrier
\section{Training Configuration}\label{app:hparams}

All nine runs share one configuration; the reward shape and the seed are the only
varied axes. Policy Qwen3.5-0.8B (post-trained hybrid); GRPO with group size
$G=5$, clip $\varepsilon=0.2$, KL penalty $\beta=10^{-3}$ with the k3 estimator,
and a group mean-and-standard-deviation baseline; learning rate $10^{-6}$,
constant, no warmup; 64 prompts per optimiser step; generation capped at 10
turns and 8192 tokens; retrieved tokens masked from the loss by message role.
Training data is the MuSiQue training split (19,968 prompts, about one epoch per
run). Retrieval is E5-base-v2 over the Wikipedia-2018 corpus through FlashRAG,
with an IVF4096-SQ8 index at top-5 for training and evaluation, and top-3 for the
Search-R1 3B reproduction. Evaluation is greedy, over 51,713 rows, with
checkpoints saved every 10 steps. Training ran on four RTX PRO 6000 Blackwell
GPUs (data-parallel degree 4, colocated vLLM), at roughly \$120 to \$140 per
full-epoch run.

Three legs ran past the seed cap of Section~\ref{sec:setup} and are truncated to
it: EM-only seed-42 to step 210 and EM-only seed-43 to step 240, neither of
which affects the reported cell, since each one's best checkpoint lies inside
the cap; and F1+format seed-43 to step 250, which reached 0.317 there against
the 0.313 we report at $\le 230$. Reinstating that 0.317 would widen the
F1+format margin over EM-only, so the truncation is the conservative choice.

\FloatBarrier

\section{Supporting Tables}\label{app:tables}

Table~\ref{tab:app-perbench} breaks the nine runs out over the seven benchmarks.
Table~\ref{tab:app-f1} reports token-F1 at each run's best-F1 checkpoint at the
matched horizon, and Table~\ref{tab:app-acc} substring accuracy at each run's
seed-horizon best-EM checkpoint; note the two use different selection rules and
different horizons, as their captions state.

\begin{table}[htbp]
\centering
\caption{Per-benchmark EM at each run's seed-horizon best checkpoint (step in the
avg-EM column). MuSiQue is the single in-distribution set; the other six are
out-of-distribution, so the generalisation claim rests on them.}
\label{tab:app-perbench}
\scriptsize
\setlength{\tabcolsep}{2.6pt}
\begin{tabular}{lrrrrrrrr}
\toprule
Run & NQ & TriQA & PopQA & HotQA & 2Wiki & MuSiQue & Bambgl & avg EM \\
\midrule
F1-only s42    & 0.301  & 0.492  & 0.319  & 0.309  & 0.287  & 0.130  & 0.352  & 0.313 @110 \\
F1-only s43    & 0.301  & 0.489  & 0.321  & 0.259  & 0.274  & 0.093  & 0.296  & 0.291 @210 \\
F1-only s44    & 0.363  & 0.531  & 0.370  & 0.359  & 0.285  & 0.161  & 0.392  & \textbf{0.352} @310 \\
F1+format s42  & 0.319  & 0.500  & 0.335  & 0.311  & 0.313  & 0.117  & 0.360  & 0.322 @100 \\
F1+format s43  & 0.335  & 0.533  & 0.352  & 0.322  & 0.296  & 0.092  & 0.264  & 0.313 @210 \\
F1+format s44  & 0.310  & 0.521  & 0.340  & 0.307  & 0.278  & 0.121  & 0.288  & 0.309 @270 \\
EM-only s42    & 0.314  & 0.492  & 0.342  & 0.306  & 0.238  & 0.103  & 0.312  & 0.301 @170 \\
EM-only s43    & 0.289  & 0.479  & 0.321  & 0.243  & 0.236  & 0.071  & 0.208  & 0.264 @120 \\
EM-only s44    & 0.343  & 0.524  & 0.383  & 0.315  & 0.283  & 0.110  & 0.264  & 0.318 @310 \\
\bottomrule
\end{tabular}
\end{table}

TriviaQA is highest everywhere (about 0.48 to 0.53) and MuSiQue lowest (about
0.07 to 0.16, the hardest set and the only in-distribution one). PopQA is where
EM-only is least disadvantaged, its short entity answers rewarding exact match.
Bamboogle carries only 125 rows, by far the smallest benchmark in the suite.

\begin{table}[htbp]
\centering
\caption{Matched horizon, average token-F1 per reward shape at each run's
best-F1 checkpoint, showing the same ordering on the dense metric.}
\label{tab:app-f1}
\begin{tabular}{lrrrrr}
\toprule
Reward shape & seed 42 & seed 43 & seed 44 & mean & range \\
\midrule
F1-only   & 0.402 & 0.369 & 0.407 & \textbf{0.393} & 0.038 \\
F1+format & 0.413 & 0.370 & 0.390 & 0.391 & 0.043 \\
EM-only   & 0.368 & 0.333 & 0.319 & 0.340 & 0.049 \\
\bottomrule
\end{tabular}
\end{table}

\begin{table}[htbp]
\centering
\caption{Substring-cover accuracy (ACC) at each run's seed-horizon best-EM
checkpoint. EM-only is worst on the most lenient metric as well.}
\label{tab:app-acc}
\begin{tabular}{lrrrr}
\toprule
Reward shape & seed 42 & seed 43 & seed 44 & mean \\
\midrule
F1-only   & 0.394 & 0.355 & 0.397 & \textbf{0.382} \\
F1+format & 0.378 & 0.359 & 0.355 & 0.364 \\
EM-only   & 0.334 & 0.293 & 0.340 & 0.323 \\
\bottomrule
\end{tabular}
\end{table}

\end{document}